\documentclass[graybox, envcountchap]{svmult}

\usepackage{type1cm}        
\usepackage{makeidx}         
\usepackage{graphicx}        
\usepackage{multicol}        
\usepackage[bottom]{footmisc}

\usepackage{newtxtext}       %
\usepackage[varvw]{newtxmath}       
\usepackage[linesnumbered,algo2e, ruled, vlined]{algorithm2e}
\usepackage{algorithm}

\usepackage[utf8]{inputenc}
\usepackage{bibunits}
\usepackage{enumerate}
\usepackage{enumitem}
\usepackage{bm}
\usepackage{booktabs}
\usepackage{multirow}
\usepackage{algpseudocode}
\usepackage{threeparttable}
\usepackage{bigstrut}
\usepackage{makecell}
\usepackage{float}
\floatstyle{plaintop}
\restylefloat{table}
\usepackage{adjustbox}
\usepackage[caption=false,font=normalsize,labelfont=sf,textfont=sf]{subfig}
\usepackage{setspace}
\usepackage{hyperref}
\hypersetup{colorlinks = true, allcolors = blue}
\usepackage{fancyhdr}
\makeindex             



\begin{document}

\title{Neural Partial Differential Equations \\
----- A Theory for Neural Architecture Evolution}
\author{Ping Guo,  Kaizhu Huang,      Zenglin Xu }
\institute{Ping Guo \at Image Processing \& Pattern Recognition Lab., Beijing Normal University,  Beijing 100875, China
\email{pguo@bnu.edu.cn}
\and Kaizhu Huang \at Data Science Research Center, Duke Kunshan University, Kunshan, Jiangsu, China 215316 \email{kaizhu.huang@dukekunshan.edu.cn}
\and Zenglin Xu \at School of Computer Science and Technology, 
Harbin Institute of Technology at ShenZhen, Peng Cheng National Lab, 
Shenzhen 510855, Guangdong, China \email{xuzenglin@hit.edu.cn}
}
%
%
\maketitle

\abstract*{In this work, we  generalize the reaction-diffusion equation in statistical physics,  Schr\"odinger equation in quantum mechanics, and Helmholtz equation in paraxial optics into the  neural   partial differential equations (NPDE), which can be considered as the  fundamental equations in the field of artificial intelligence research.
We take the finite difference method  to discretize  NPDEs for finding the numerical solution. Moreover, the basic building blocks of  deep neural network  architectures, including multi-layer perceptron, convolutional neural networks, and recurrent neural networks,  are generated.  The learning strategies, such as Adaptive moment estimation, L-BFGS, pseudoinverse learning  algorithms, and partial differential equation constrained optimization, are also developed. 
This work is  of significance in that it presents a clear view of interpretable deep neural networks physically, enables high possibility to be applied to analog computing device design, and paves the road to physical artificial intelligence.}

\abstract{In this work, we  generalize the reaction-diffusion equation in statistical physics,  Schr\"odinger equation in quantum mechanics, and Helmholtz equation in paraxial optics into the  neural   partial differential equations (NPDE), which can be considered as the  fundamental equations in the field of artificial intelligence research.
We take the finite difference method  to discretize  NPDEs for finding the numerical solution. Moreover, the basic building blocks of  deep neural network  architectures, including multi-layer perceptron, convolutional neural networks, and recurrent neural networks,  are generated.  The learning strategies, such as Adaptive moment estimation, L-BFGS, pseudoinverse learning  algorithms, and partial differential equation constrained optimization, are also developed. 
This work is  of significance in that it presents a clear view of interpretable deep neural networks physically, enables high possibility to be applied to analog computing device design, and paves the road to physical artificial intelligence.}


\section{Introduction}
\label{sec1}

Over the years, artificial intelligence (AI) or more specifically digital artificial intelligence has made rapid progress. Recently, researchers have shifted their attention to  physical artificial intelligence (PAI),  believing that PAI could be the most potential field in robot field in the next decades~\cite{Miriyev2020}.  While theories and methodologies from physics have been  boosting artificial intelligence methods, we believe they are able to  pave the road to PAI.
 To this end,  a solid foundation should be set out from theoretical physics in advance in order to develop PAI effectively.

It is well-known that in physics, theoretical work is said to be from first principles,  if it  does not make assumptions.  Moreover, in physics,  the principle of least action (PLA) is  a very elegant  and important discovery of mechanics~\cite{BASSET1903}.
The PLA can be used to derive Newtonian, Lagrangian and Hamiltonian equations of motion, and even general relativity.   
  The physicist  Richard Feynman, demonstrated how the PLA can  be used in quantum mechanics with path integral calculations~\cite{Feynman2005}.   It is also well known that most of the fundamental laws in Physics  are described with partial differential equations (PDE).  As Stephen Wolfram said, 
 ``{\em Indeed, it turns out that almost all the traditional
mathematical models that have been used in physics and other areas of
science are ultimately based on partial differential equations}''~\cite{Wolfram2002}.
  
Now the question is then what is the first principle of AI?   Are there any fundamental equations for the AI field?  In our viewpoint, if  there exists the first principle for AI,  we believe it should be  the PLA~\cite{firstprinciple2020}.  Also, as the problem asked by Yann LeCun in  IJCAI 2018, ``{\em What is the equivalent of  `Thermodynamics' in intelligence? }" \cite{LeCun2018jicai},
  this has inspired us to seek the simple rule to build a world model, or, to establish the  fundamental equations for AI with physics laws inspired methods. 

In this work, we  generalize nonlinear parabolic partial differential equations (PPDE) into neural partial differential equations (NPDE),  which can be considered as the  fundamental equations in the  AI research field. We also show that 
the reaction-diffusion equation in statistical physics,  Schr\"odinger equation in quantum mechanics, Helmholtz equation in paraxial optics, are all examples of PPDEs. With the finite element method (FEM)  and the finite difference method (FDM), we discrete  NPDEs to generate neural network architecture. And we believe that NPDEs can be used to describe PAI systems, such as the synergetic learning systems.

 \section{Background} 
In this section, we briefly review the synergetic learning systems (SLS)~\cite{Synergetic2019} followed by a simple introduction to Turing's equations. 

 \subsection{SLS Theory}
 Our proposed  SLS is an information processing system based on the PLA, and  the dynamics of the SLS should be described by differential dynamic equations. 
It is well-known  that when generalizing the equation to  multi-component systems, a variation of the ``free energy'' $\mathcal{F}$ can be obtained:
\begin{eqnarray}
\label{generalRDE}
\frac{\partial \Psi }{\partial t}=\nabla \cdot (D(\Psi )\nabla\Psi )+\nabla f(\Psi)+g(\Psi), 
\end{eqnarray}
where $\Psi (x,\theta ,t)=\left \{ \psi _{1},\psi _{2},\cdots,\psi _{m}  \right \}$, $D(\Psi)$ is the diffusion matrix, and 
$\nabla \Psi=\left ( \nabla  \psi_1, \nabla  \psi_2, \cdots, \nabla  \psi_m\right )$,
$\nabla \psi_i=\left ( \frac{\partial \psi_i}{\partial x_{1}}\vec{e_{1}}+\frac{\partial \psi_i}{\partial x_{2}}\vec{e_{2}}+\cdots \cdots+ \frac{\partial \psi_i}{\partial x_{n}}\vec{e_{n}}\right )$. 
$\nabla f(\Psi)$ is called the convection vector and $g(\Psi)$ is the reaction vector\cite{Qixiao16}, and $\vec{e_{i}}$ stands for unit basis vector.

Therefore, the dynamics of the SLS are described by the reaction-diffusion equations (RDE). In statistical physics, the dissipative system theory is a theoretical description of the self-organization of non-equilibrium systems. In addition,  RDEs are utilized to model the systems~\cite{Willems1972}.  If the SLS is designed as a differential dynamical system and attention is paid to the attractor subnetwork~\cite{DomnisoruMembrane22}, RDEs can also be used to describe this kind system. 

\subsection{Turing's Equations}
Turing's  equations  are  two-component RDEs~\cite{Turing1952},  and  the mechanisms for pattern formation~\cite{Pearson1993}.
Letting $\psi _{1}=U $ and $\psi _{2}=V$,  the mathematical expression of  Turing's  equations is 
\begin{eqnarray}
\label{turing-equations}
 \frac{\partial U}{\partial t} &=& D_u \nabla ^2 U + f (U, V) \nonumber \\ 
 \frac{\partial V}{\partial t} &=& D_v \nabla ^2 V + g (U, V).
\end{eqnarray}

In the literature~\cite{Self-Organization26},  Prigogine proposed the theory of dissipative systems, and founded a new discipline called  non-equilibrium statistical mechanics, and  RDEs are then utilized to model the dissipative structures.

\section{Neural Partial Differential Equations}
\label{sec-NPDEs}
In this section, we will  discuss various equations from the perspective of PPDEs. We then present a general form for NPDEs.

As a family of semi-linear PPDEs, RDEs are widely used in statistical physics. We believe that nonlinear PDEs can be utilized to describe an AI system, and they can be considered as  fundamental equations for neural  systems.
In the following, we will present a general form of NPDEs.

Taking matrix-valued function ${A}(U({x}, t))$, ${B}(U({x}, t))$ and ${C}(U({x}, t))$,  we can define the  elliptic operator $\mathcal{O}_L $ ~\cite{Wells2008Elliptic}:
\begin{eqnarray}
\label{ellopticopra}
 \mathcal{O}_L U=&-&\nabla \cdot (A ( U(x) )\nabla U) \nonumber\\
 &+&B( U(x) ) \nabla U + C(U(x))  U.
 \end{eqnarray}

This is a more general second-order divergence form for the nonlinear elliptic differential operator.  When the matrix-valued function is only a function of $x$, the operator  is a linear elliptic differential operator. Especially,  if  the matrix-valued function $A (x)$ 
 is symmetric and positive definite for every $x$,  it is said to be a uniformly elliptic operator.

It should be noted that this  elliptic operator form is the  same as the right side of Eq.~(\ref{generalRDE}).  The matrix   $A(U)$  is the diffusion matrix,  the second term in the right side of Eq.~(\ref{ellopticopra}) is  the convection vector, and the third term is the reaction vector.

In mathematics,  the Laplace operator is expressed as:
\begin{equation}
 \nabla^2  U =: \triangle U  = \sum _i {\partial^2 \over \partial x_{i}^{2}} U.
\end{equation}
In the theory of PDEs, elliptic operators are differential operators that generalize the Laplace operator, since 
the Laplace operator is obtained by taking $A = I$.  

In mathematics, a PDE is an equation which imposes relations between the various partial derivatives of a multi-variable function. Furthermore, second-order linear PDEs are classified as either elliptic, hyperbolic, or parabolic. As for nonlinear PDEs,  no mathematics can be strictly  defined whether the  nonlinear PDEs belong to  parabolic,   hyperbolic, or elliptic. In the following we just take several  nonlinear PDEs besides RDEs.

\subsection{Fisher's Equations}
\label{Fisher-EQ}
The Fisher's equation,  also known as the Kolmogorov-Petrovsky-Piskunov (KPP) equation,  or Fisher--KPP equation~\cite{FisherEQ1995}, takes the following mathematical expression
\begin{equation}
{\frac {\partial U}{\partial t}}= D{\frac {\partial ^{2}U}{\partial x^{2}}} + rU(1-U).
\end{equation}
The Fisher's equation belongs to the class of RDEs.  In fact, it has the inhomogeneous term and is deemed as one of the simplest semi-linear RDEs. 

\subsection{Heat Equations}

The heat equation is a certain PDE in mathematical  physics.  When setting  the diffusion
matrix $D=I$,  it has the form~\cite{heateq1964}
\begin{equation}
\frac {\partial u}{\partial t}={\frac {\partial ^{2}u}{\partial x_{1}^{2}}}+\cdots +{\frac {\partial ^{2}u}{\partial x_{n}^{2}}},
\end{equation}
where $(x_1, \dots., x_n, t)$ denotes a general point of the domain.  

When one studies the thermal-optic effect in the nonlinear optical medium, the  heat equation has the form
  \cite{Bigot1987}\cite{Guo1993Heat}\cite{Guo1993HeatDyna}:
\begin{equation}
\label{heat-conduct}
\frac {\partial T}{\partial t}=\frac {\partial }{\partial z}\left (  \frac{\kappa}{\rho c} \frac {\partial T}{\partial z} \right) +\frac{\alpha} {\rho c} I (z, t), 
\end{equation}
where $T$ is temperature,  $\kappa, \rho, c $,  and $ I (z, t)$, are heat conductive coefficient, medium density,  medium specific heat capacity and light intensity, respectively. 

\subsection{Schr\"odinger Equations}
The Schr\"odinger equation is a linear PDE that governs the wave function of a quantum-mechanical system~\cite{quantumMe2007}. The most general form is the time-dependent Schr\"odinger equation:
\begin{equation}
i\hbar {\frac {d}{dt}}\vert \Psi (t)\rangle ={\hat {H}}\vert \Psi (t)\rangle, 
\end{equation}
where  $i$ denotes the imaginary index, $\hbar$ denotes the reduced Planck constant, $\Psi $ is the state vector of the quantum system,  $t$ is time, and $\hat {H}$ is an observable  Hamiltonian operator.

The Schr\"odinger equation  does have the same form as the heat equation, when we perform  Wick rotation\cite{Wick1954}. For example,  one-dimensional Schrödinger equation for a free particle is
\begin{equation} 
\label{eq:se1} 
i\hbar\frac{\partial\psi(x, t)}{\partial t}=-\frac{\hbar^2}{2m}\frac{\partial^2\psi(x,t)}{\partial x^2}. 
\end{equation}

If one takes the Wick rotation $ \tau=it$ , the Schr\"odinger equation \eqref{eq:se1} turns into
\begin{equation} 
\label{eq:hheq} 
\frac{\partial\phi(x,\tau)}{\partial\tau}=\frac{\hbar}{2m}\frac{\partial^2\phi(x,\tau)}{\partial x^2}, 
\end{equation}
where $\phi(x,\tau)=\psi\left(x, {\tau}/{i}\right) $. Eq. ~\eqref{eq:hheq} is a homogeneous heat equation with the diffusion coefficient 
$\alpha^2=\frac{\hbar}{2m}$. Conversely,  if we apply the Wick rotation $\tau=-it$ to the one-dimensional homogeneous heat equation, i.e.
\begin{equation} 
\label{eq:hhe2} 
\frac{\partial u(x,t)}{\partial t}=\alpha^2\frac{\partial^2 u(x,t)}{\partial x^2},
\end{equation}
then the resulting equation is
\begin{equation} 
\label{eq:se2} 
i\hbar\frac{\partial w(x,\tau)}{\partial \tau}=-\alpha^2\hbar\frac{\partial^2 w(x,\tau)}{\partial x^2}, 
\end{equation}
where $w(x,\tau)=u\left(x,-{\tau}/{i}\right)$. Eq.~\eqref{eq:se2} is the  Schr\"odinger equation for a free particle with 
$m={\hbar}/({2\alpha^2})$. 

This example indeed shows an intriguing relationship between  Schr\"odinger equation and heat equation.

Also, by Wick rotation  $\tau=it$, we have the  Schr\"odinger equation for a particle in a potential $V$:
$$
\hbar \frac{\partial\psi}{\partial t}=\frac{\hbar ^2}{2m}\nabla^2 \psi -V\psi.
$$
This equation has the same form as the heat diffusion equation.

 \subsection{Paraxial Helmholtz Equations}
 In mathematics, the eigenvalue problem for the Laplace operator is known as the Helmholtz equation. It corresponds to the linear PDE:

$$
\nabla ^{2}U=-k^{2}U,
$$
where $\nabla ^{2}$ is the Laplace operator, $k^2$ is the eigenvalue, and $U$ is the eigenfunction. When the equation is applied to waves, $k$ is known as the wave number. The Helmholtz equation has a variety of applications in physics, including the wave equation and the diffusion equation.

A wave is said to be paraxial if its wavefront normals are paraxial rays. In paraxial Optics, parabolic approximation, also called slowly varying envelope approximation (SVEA)~\cite{Svelto1974}, is utilized. 
In the paraxial approximation of the Helmholtz equation~\cite{photonics2019}, the complex amplitude $U$ is expressed as
\begin{equation}
\label{eq:phe1}
 U(\mathbf{r}) = u(\mathbf{r}) e^{ikz}, 
\end{equation}
where $u$ represents the complex-valued amplitude  modulating the sinusoidal plane wave represented by the exponential factor. Then under a suitable assumption, $u$ approximately solves
\begin{equation}
\label{eq:phe2}
\nabla_{T}^2 u + 2ik\frac{\partial u}{\partial z}  = 0,
\end{equation}

where $\nabla_{T}^2 =: \partial ^{2}/\partial x^{2}+\partial ^{2}/\partial y^{2} $ is the transverse Laplacian operator.

This is a PPDE,  representing waves propagating in directions significantly different from the $z$-direction. In addition, it has important applications in  optics science, by offering solutions that describe the propagation of electromagnetic waves in the form of either paraboloidal waves~\cite{Guo1990AOS} or Gaussian beams~\cite{Guo1990}\cite{Guo1999Dynamics}. 

 \subsection{General NPDEs}
In the above, we present several PPDEs, including backward PPDEs. It is noted that PPDEs can also be nonlinear. For example, Fisher's equation is a nonlinear PDE that includes the same diffusion term as the heat equation but incorporates a linear growth term and a nonlinear decay term.

Now we formulate a general form of  NPDEs named as nonlinear PPDEs
\begin{equation}
\label{neuralPDE}
 \frac{\partial \Psi }{\partial t}=\mathcal{F}\left [ x, t,  \nabla, \nabla ^2, A ( \Psi ), B( \Psi ),  C (\Psi), \Psi \right ].
 \end{equation}
 
With the quasi-linear approximation, Eq.~(\ref{neuralPDE}) will be substituted with
\begin{eqnarray}
\label{neuralPDE1}
 \frac{\partial \Psi }{\partial t}&=&\mathcal{O}_L \Psi  \\
\mathcal{O}_L \Psi  &=&\nabla \cdot (A ( \Psi )\nabla \Psi)+ B( \Psi ) \nabla \Psi + C(\Psi) \nonumber.
\end{eqnarray}
 Here the quasi-linear approximation means that matrices $A( \Psi ), B( \Psi )$ are the linear functions of $\Psi$,  while $C(\Psi)$ can be the nonlinear function of  $\Psi$, for example, $C(\Psi)= 1/(1+\exp(-\Psi))$.

\section{Generating Neural Architectures}
In this section, we introduce how to  generate different neural architectures from NPDEs. We will discuss convolutional neural network (CNN), Full-connection-DNN, and recurrent neural network  (RNN) in turn.

\subsection{Two-model  SLS}
\begin{figure}[htbp]
\centerline{\includegraphics[width=7cm]{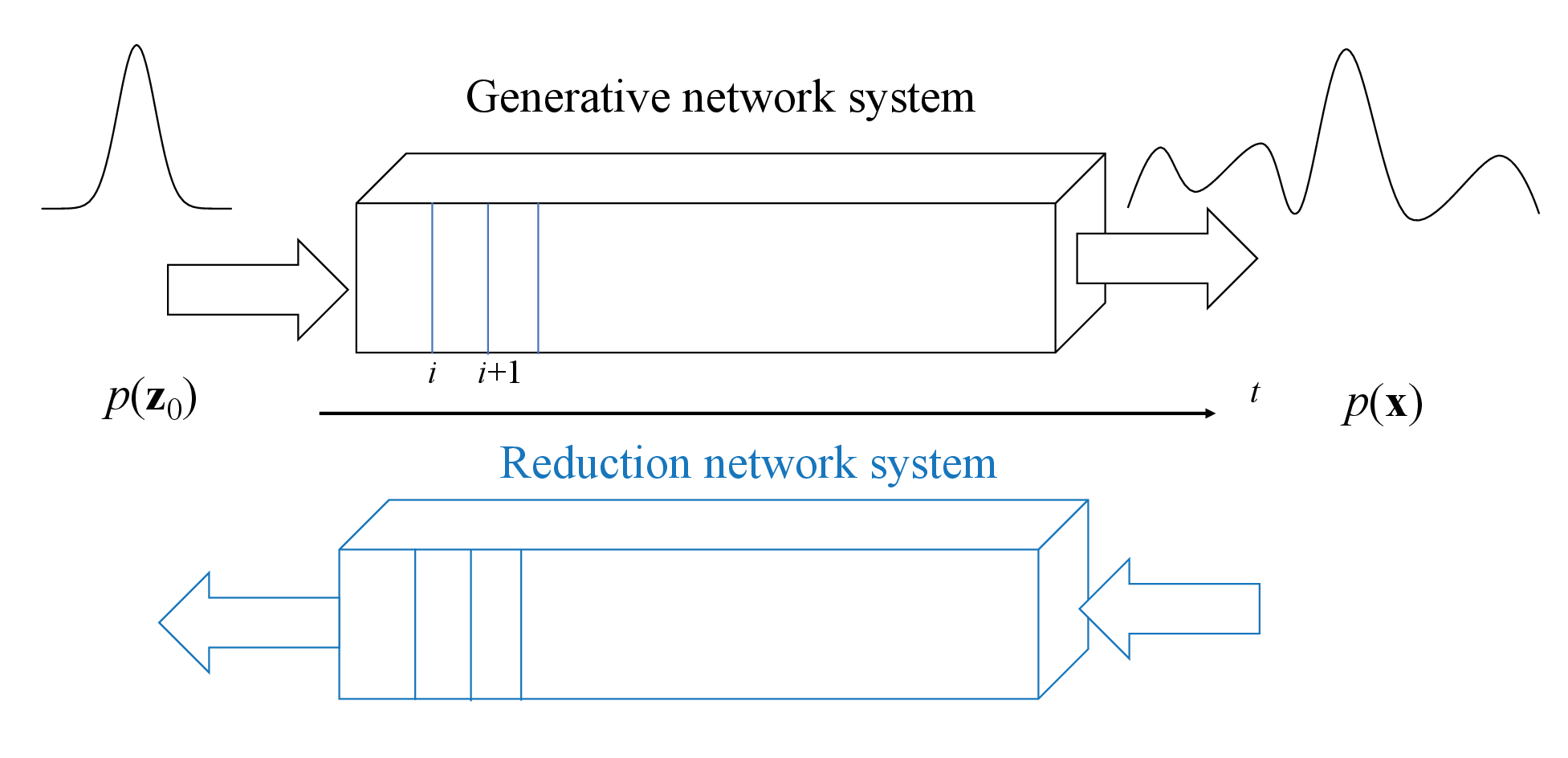}}
\caption{An SLS is constructed with  two-model case: system generative model (System A) and system reduction model (System B).}
\label{two-model-SYS}
\end{figure}

 We discuss two-model SLS in~\cite{Synergetic2019b}\cite{SynergeticIII2020}. Fig.~(\ref{two-model-SYS}) shows our designed systems.

The generative model is called  System A, and RDE is adopted  to describe the system evolution procedure. We also introduce an
auxiliary model, defined as the reduction model (System B), to solve the inverse procedure problems.
 
\subsection{Finite Difference Methods}
At present, most studies on PDEs in mathematics exploit FDM~\cite{FerzigerFinite28} to obtain numerical solutions~\cite{Grossmann2007}, along with FEM~\cite{Zienkiewicz2013}.  FEM, as one  popular numerical method, aims to solve various differential equations arising in many areas such as fluid flow and heat transfer by decomposing a large system into smaller and simpler ones called finite elements.  FEM usually results in a system of algebraic equations. Along with FEM, FDM presents one important numerical technique for solving differential equations but by mainly approximating derivatives with finite differences. FDM transforms nonlinear ODE or PDE, into a system of linear equations that can be solved efficiently.

Before introducing our method, we first describe the preliminary of FDMs for the discrete Laplacian.
For nonlinear PDEs,  the numerical solution method  is usually used~\cite{numericalRDE1}\cite{numericalRDE}.

In the one-dimension case, the discrete Laplace operator is approximated as
\begin{equation}
\Delta u(x)\approx {\frac {u(x-h)-2u(x)+u(x+h)}{h^{2}}}=:\Delta _{h}u(x)\, .
\end{equation}
This approximation is usually expressed via the following 3-point stencil
\begin{equation}
\label{1Dlaplace}
 \frac {1}{h^{2}}{\begin{bmatrix}1 & -2 & 1\end{bmatrix}}.
\end{equation}

 As we know that the widely used  Laplace operator is Laplace of Gaussian in image processing, the discrete Laplace operator is a 2D filter~\cite{GuoYZ2015image}:   
\begin{equation}
\nabla_{T}^2 ={\begin{bmatrix}0&1&0\\1&-4&1\\0&1&0\end{bmatrix}}.
\end{equation}
$\nabla_{T}^2$  corresponds to the (five-point stencil) finite-difference formula.   When  the nine-point stencil is used, it includes the diagonals:
\begin{equation}
\label{transverseDLO}
\nabla_{T}^2 ={\begin{bmatrix}0.25&0.5&0.25\\0.5&-3&0.5\\0.25&0.5&0.25\end{bmatrix}}.
\end{equation}
The stencil  represents a symmetric matrix. If  an equidistant grid for a finite element unit is utilized, one will get a Toeplitz matrix.

Next, we will present three most popular DNN building blocks.  DNN can be built with simple  stacked blocks in a combined cascade/parallel manner. In physics, when one studies the field distribution in the layered-wise structure medium, PDEs will be used to find the system solution. The layer-wise structure medium model shares certain relationship with the DNN model. For instance, they enjoy the same layer-wise structure. This implies that the DNN structure is discretized from the layer-wise structure medium physical model.
   
\subsection{Convolutional Neural Networks}
\label{CNN-discribe}
 Suppose we use one component RDE to describe the System A in SLS as presented in Fig.~(\ref{two-model-SYS})
 \begin{equation}
 \label{CNNgen}
 \frac{\partial u}{\partial t} = \underbar{D} \triangle u + f (u, t). 
\end{equation}
 In the following, we will first discuss one dimensional signal processing  with the CNN problem.

 \subsubsection{1D CNN}
For the one dimensional case, operator $\triangle = {\partial ^2 }/{\partial  x ^2} $. For extremely deep and wide neural  networks, $x$ axis stands for  the width direction of a layer in the network.   As discussed in~\cite{Guo2020archEvolve},  the ResNet-like
MLP is developed.  Now we assume that $\underbar{D}  = A(x, t)$ and $f(u, t)= \sigma (u, t)$ is an activation function.  
\begin{equation}
 \label{FE2}
\frac{\partial u}{\partial t}=A(x, t)\frac{\partial ^2 u}{\partial  x ^2} +\sigma (u, t).
\end{equation}
  
Taking Euler's method to the above PDE, we have
\begin{equation}
 \label{FE3}
\frac{u(t+k)-u(t)}{k}=A(x, t)\frac{\partial ^2 u}{\partial x ^2 } +\sigma (u, t).
\end{equation}
 When taking step $k=1$, we get
 \begin{equation}
 \label{FE4}
u(t+1)=u(t)+A(x, t)\frac{\partial ^2 u}{\partial  x ^2} +\sigma (u, t).
\end{equation} 

Almost all the previous works deem $\mathcal{F}(u, \triangle u, t, \sigma )=  A(x, t)\frac{\partial ^2 u}{\partial x ^2 } +\sigma (u, t )$
as a black box, and only consider the network depth direction $t$ variation~\cite{ChenRBD2018ode}\cite{LongLD2019PDEnet} \cite{LuZLD2018icml}. We now open this black box, to investigate the NPDE in the $x$ (width) direction  case, 
 explain why this RDE is  an  NPDE, and introduce how it can  generate neural architectures.


  
Here we describe the physical view of our model.  Imagine  that the information is expressed with variable $u(x, t)$, which propagates through a 2-dimensional virtual medium, e.g., a thin film. This  thin film has no thickness, and only has width ($x$) and length ($z$).  The information along $z$ direction propagates with a constant speed\footnote{Strictly speaking, light speed is not a constant in the nonlinear medium. Here we just neglect the nonlinear effect.} $v$,  when adopting traveling wave solution, we let $\tau = z- v t$.  
 
 $$
 \frac{\partial u}{\partial t} = \frac{\partial u}{\partial \tau}\frac{\partial \tau}{\partial t} = - v \frac{\partial u}{\partial \tau}= - v \frac{\partial u}{\partial z} .
  $$ 
  
Therefore, using symbol $z$ or $t$  only leads to a constant $v$ difference in discretizing PDE for time difference.  Recall our infinite width and  infinite depth neural network model~\cite{Guo2020archEvolve},  where $x$  stands for network's width direction,  and $t$ (or $z$) means network's depth direction.

\begin{figure}[htbp]
\centerline{\includegraphics[width=8cm]{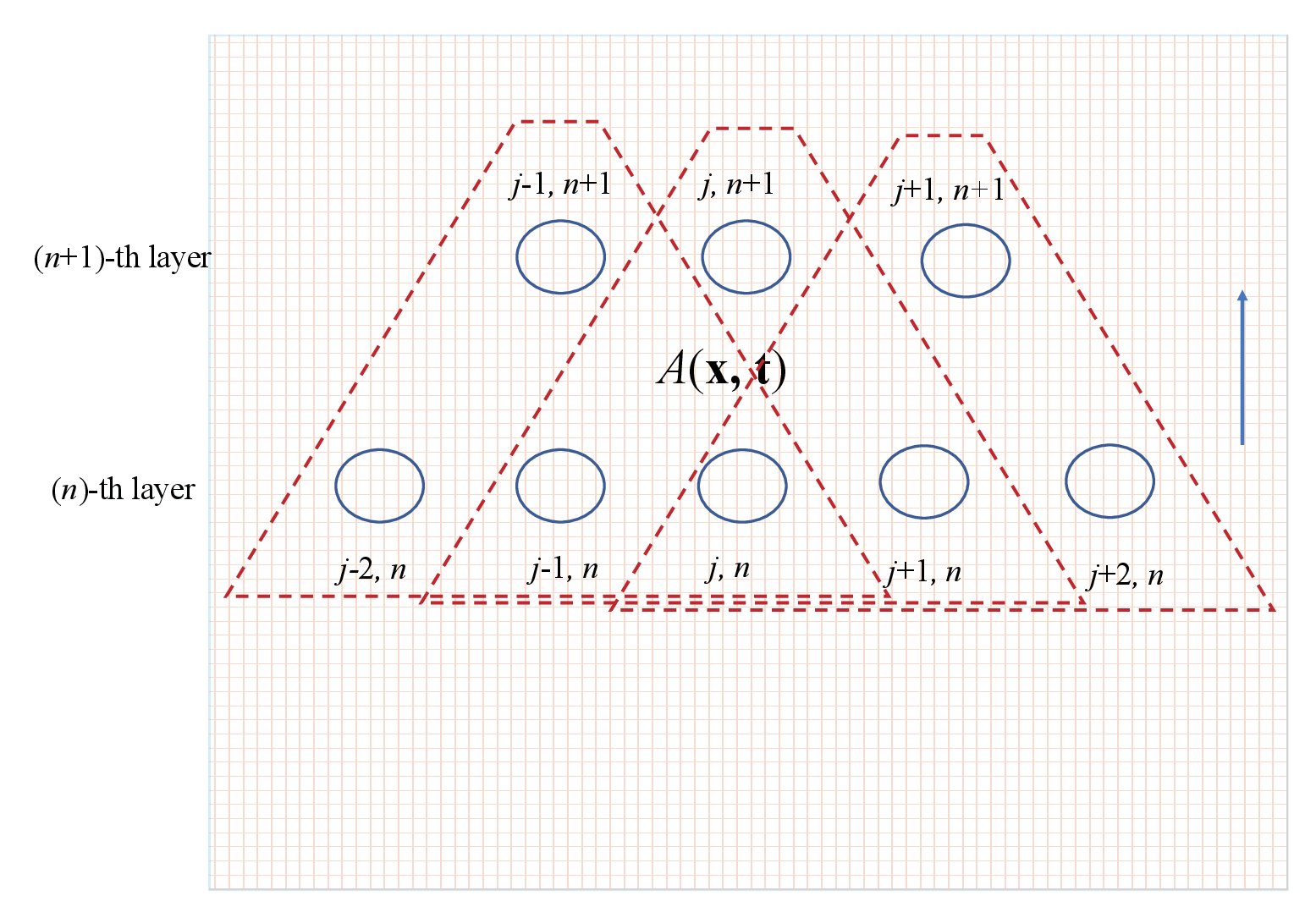}}
\caption{Explanation of the elliptic operator for neural network based   physics model described with RDEs.}
\label{fig-neuralnet}
\end{figure}

Let us give our interpretable NPDEs with Fig.~(\ref{fig-neuralnet}).
 
In  Fig.~(\ref{fig-neuralnet}), $n$ and $j$ are explained as network layer index ($z$), and network neuron position ($x$) index,   respectively.  Variables, which are explained as the network's depth and width variable, respectively,  is the function of $x$. $A(x, t)$  is  related  to the properties  of  the medium model which can be considered as diffusion matrix $\underbar {D}$ in RDEs.
 This 2D virtual medium is inhomogeneous medium when we assume $A(x, t)$  as the function of $x$ and $t$. This resembles the weight matrix in DNN where each layer is not the same, since  $A(x, t)$ is dependent on $t$ (or $z$) variables.
 
 If we combine Eq.~(\ref{1Dlaplace}),  the elliptic operator in Eq.~(\ref{FE4}) (the operator part of the second term) can be written in the form:
\begin{equation}
\label{3elliptic-operator}
 \frac {1}{h^{2}}{\begin{bmatrix}A(x_{j-1}, t_n) & -2A(x_j, t_n) & A(x_{j+1}, t_n)\end{bmatrix}}\;.
\end{equation}

Since the modeled medium is assumed inhomogeneous medium,  $A(x, t)$   is unknown and  may take any values. If we intend to take weight share trick as in CNN,  $A(x, t)$ should be assumed as a special function.
Hence we let this stencil 
\begin{equation}
\label{weight-matrix}
 {\begin{bmatrix}W_{j-1, n} &W_{j, n} & W_{j+1, n}\end{bmatrix}}
\end{equation}
as a weight matrix to be learnt in CNN. The second term in Eq.~( \ref{FE4}) becomes
\begin{equation*}
\label{convolution1}
 {\begin{pmatrix}W_{j-1} &W_{j} & W_{j+1}\end{pmatrix}_n}
  \begin{pmatrix}
 u_{j-1}      \\
 u_{j}     \\ 
 u_{j+1}
\end{pmatrix}_n= \sum_{i=-1}^{i=+1} W_{j+i}u_{j+i}=a_{j, n+1}.
\end{equation*}

We explain  $a_{j, n+1}$  to be  the $j$-th neuron input at layer $n+1$.   Similarly,  from Fig.~(\ref{fig-neuralnet}), we have
$$
a_{j+1, n+1}= \sum_{i=-1}^{i=+1} W_{j+1+i}u_{j+1+i},
$$   
$ a_{j-1, n+1}= \sum_{i=-1}^{i=+1} W_{j-1+i}u_{j-1+i},$   and so on.   

 Apparently, the  elliptic operator plays  a convolutional operation role. In this example, the 1D convolutional  kernel size is 3.  Also, after  the convolutional operation is implemented, $t$ slice is increased by a small amount of quality, say $0.8k$.    

The third term in Eq.~(\ref{FE4}) will be taken as pooling and/or activation nonlinear operation. For Fisher's equation,  we observe that $\sigma (u, t)= r u(1-u)$.  When $u(r x)=  {1}/{(1+e^{-r x})}$  is a sigmoid function,  $\sigma (u, t) =\partial u / \partial x $. (Suppose that $x, t$ are separable).

It is noted that information propagates  at most with the light speed $c$ because of physics law restriction. When we calculate pooling and/or activate operation, we use the function $u(t+0.8k)$, meaning that  the pooling or activate operation is after the completion of convolution operation.  A general expression can be $F(t_1) + C(t_{1+\delta}) \sim C( F(t_{1+\delta})) $, where $\delta$ stands for a very small quality,  $F(\cdot)$ and $C(\cdot)$ are some functionals, and $C(\cdot)$ may play the role of activation function in neural networks. In this way, the nonlinear PDE is approximated with the quasi-linear PDE.

Here we obtain the basic building block of 1D CNN from NPDE by applying FDM.  When we assign $A(x, t)=k_0$,  $\sigma (u, t) =  k_1 u(1-u)$, Eq.~(\ref{FE2}) can be reduced into Fisher's equation, 

\begin{equation}
 \label{FE1}
\frac{\partial u}{\partial t}=k_0\triangle u + k_1 u(1-u).
\end{equation}

\subsubsection{2D CNN}

\begin{figure}[htbp]
\centerline{\includegraphics[width=8cm]{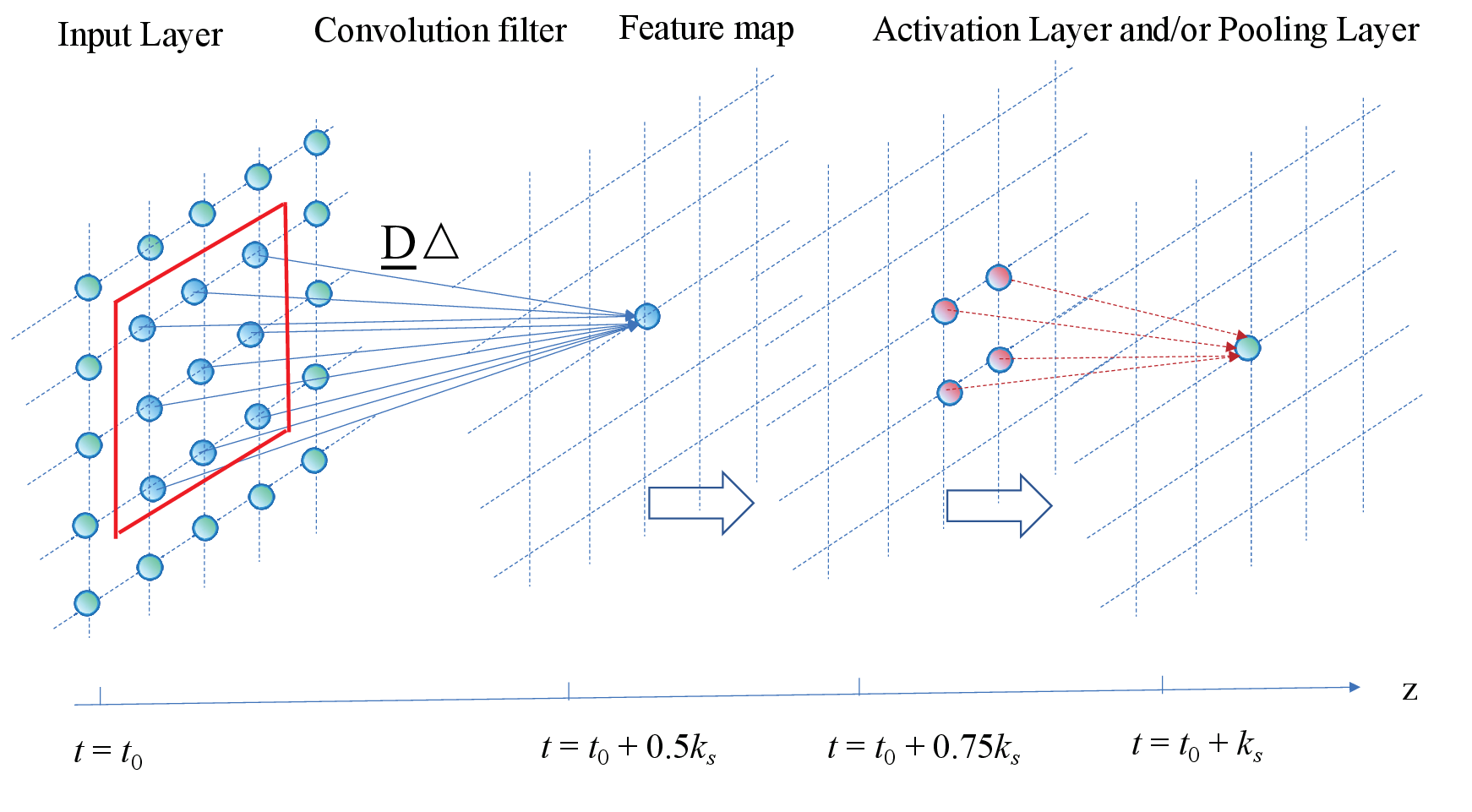}}
\caption{Discretized diffusion equation as CNN. The time slice is
equivalent to the layered structure along $z$ direction.}
\label{fig-CNN}
\end{figure}
  
 Our  idea is that, by incorporating diffusion matrix $\underbar {D}$ with  discrete transverse Laplace operator expression 
 Eq.~(\ref {transverseDLO}), and the combined filter is assigned as a learnable filter, the elements in this filer will be learned like the convolutional kernel in CNN.  In fact, the following filter is a special case of the Elliptic operator.
\begin{equation}
\label{2Dfilter}
\underbar{D}\triangle = 
{\begin{bmatrix} 
W _{1,1}& W_{1,2} & W_{1,3}\\
W_{2,1}& W_{2,2} & W_{2,3}\\
W_{3,1}& W_{3,2} & W_{3,3}
\end{bmatrix}}. 
\end{equation}
Here we can regard this filter as a second order derivative operator.  

With the above analysis,  it would be very easy to discrete NPDEs to the ResNet structure. While  it is widely-known that ResNet can be considered as ODE net,  our analysis indicates how ResNet can be obtained by  discretizing  RDEs.  

Eq.~(\ref{2Dfilter}) shows the 9-point stencil, which is equivalent to $3 \times 3$ convolutional kernel.  A visualized expression can be seen in Fig.~(\ref{fig-CNN}). Apparently, it will proceed  two periods  in the right side part of Eq.~(\ref {FE3}) when the left side steps one in  ResNet. In other words, $k_s=1/2 \, k$  in Fig.~(\ref{fig-CNN}) for ResNet.

\begin{figure}[htbp]
\centerline{\includegraphics[width=8cm]{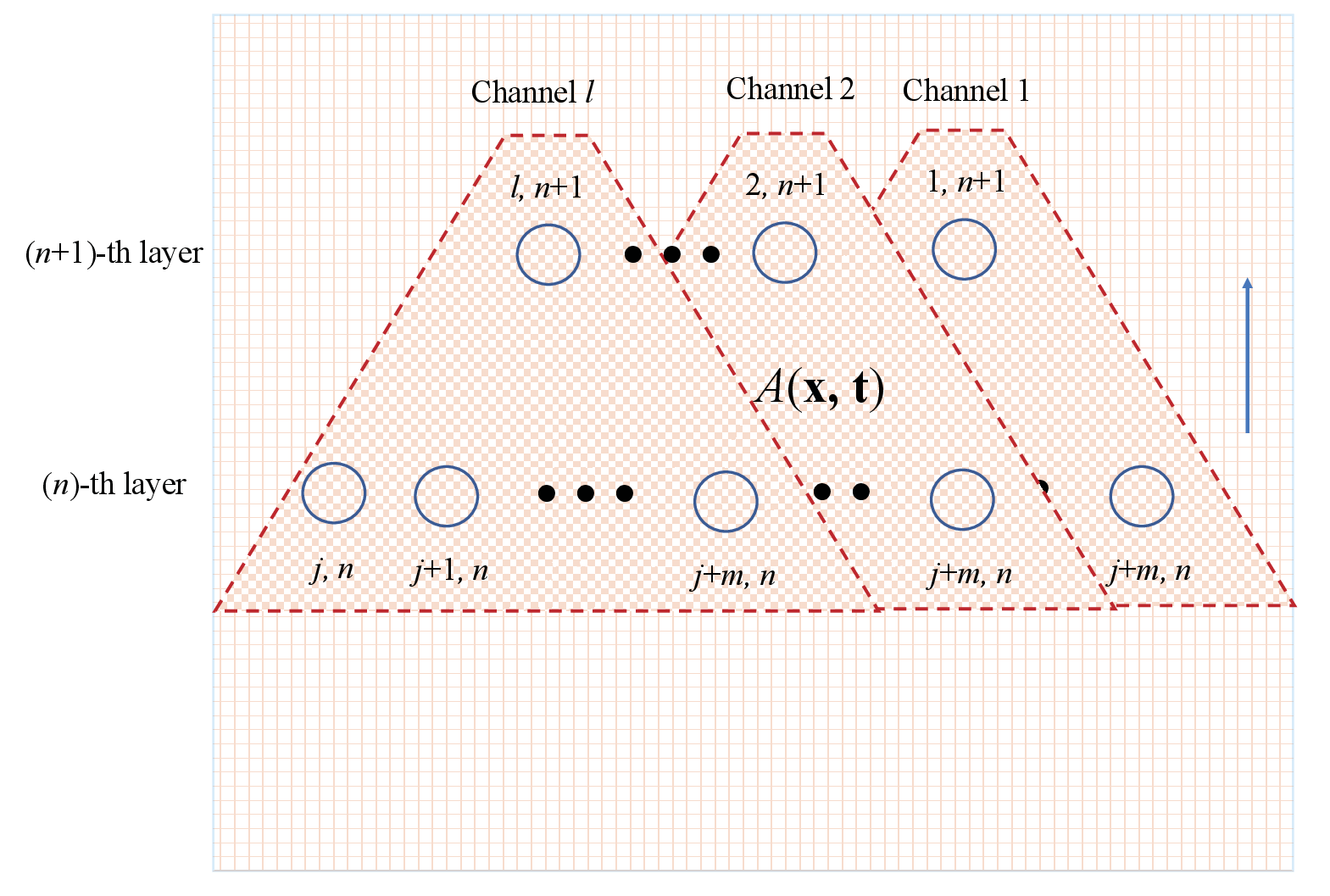}}
\caption{Multi-channel with full size convolutional kernel is equivalent to  an MLP network structure.}
\label{fig-channel}
\end{figure}
  
\subsection{Full Connection--DNN (MLP)}
 \label{subsec:mlp}
  MLP is a well-studied structure of DNN. As proved in~\cite{Guo2020archEvolve},  the full connected MLP is equivalent to the full size kernel convolution in addition with the multi-channel operation.  The channel number is the desired neuron number in the ${(n+1)\text{-th}}$ layer.  Fig.~(\ref{fig-channel})  illustrates this relationship between the full link layer and the full size kernel with a multi-channel design architecture.
  
For example, suppose the $n$-th layer and $(n+1)$-th layer have $m$ and $l$ neurons, respectively. Then the information at channel $i$ propagated to reach the activation function input is
\begin{equation}
\label{channel-1}
 a_{i, n+1}= \sum_{j=1}^{m} W_{j, i}u_{j, n}.
\end{equation}
We need to design $l$ channels to get $l$ neurons  at  the  $(n+1)$-th layer. 

As discussed in~\cite{Guo2020archEvolve}, random vector functional link network (RVFL) can evolve to a ResNet-like MLP with our design. The  output of the ResNet-like MLP's  layer $ L_{l+1} $  is\footnote{The symbol $\mathbf {x}$ in Eqs.~(\ref{resnet2mlp0}) and (\ref{resnet2mlp}) has the same meaning as $u$ in this chapter.}

 \begin{equation}
 \label{resnet2mlp0}
 \mathcal{F}(\mathbf {x}_l, \mathcal{W}_l) + \mathbf {x}_l.
 \end{equation}
This is the same as  Eq.~(3)  in~\cite{eccvHeZRS2016ResNet2}. 
 \begin{equation}
 \label{resnet2mlp}
\mathbf {x} _{l+1} = \mathbf {x}_l +\mathcal{F}(\mathbf {x}_l, \mathcal{W}_l).
 \end{equation}
It is easily verified that this equation is the discrete ODE by Euler's  method.  In comparison, in this work, we consider not only  the ODE, but also PDE.  Eq.~(\ref {channel-1}) is just the second term of  Eq.~(\ref{FE4}).  When it is fed into an activation function $\sigma (u, t)$ of hidden neuron $j$,   
 $\mathcal{F}(\mathbf {x}_l, \mathcal{W}_l)$ will be figured out.
 Also  quasi-linear approximation is utilized to solve nonlinear PDE.
 
In the above, we present the  full connection NN such as MLP. The weight layer can  also be regarded as  full size convolution kernel combined with multi-channel operator~\cite{Guo2020archEvolve}, which is naturally transformed to NPDEs  accordingly.

In addition,  restricted Boltzmann machine (RBM) was proposed by Hinton {\em et al}. in 2006~\cite{Hinton2006}, which can be regarded as a generative stochastic neural network. In fact, the topological structure of RBM  is essentially the same as a full link single layer feed forward neural network where the connecting weights and neurons are stochastic.  And RBM can be generated with stochastic partial differential equations, details are presented (see Appendices).

 \subsection{Recurrent Neural Network}
 \label{RNNsubsection}
RNNs are widely used in  temporal sequence processing. They  can exhibit the temporal dynamic behavior~\cite{Lipton2015RNN}.
Hopfield  network is a special kind of RNN,  while the modern Hopfield  network proposed by Ramsauer {\em et al}.~\cite{HopfieldNAYN2020} may have broad applications compared with attention mechanics~\cite{nipsVaswaniSPUJGKP2017}.

\begin{figure}[htbp]
\centerline{\includegraphics[width=8cm]{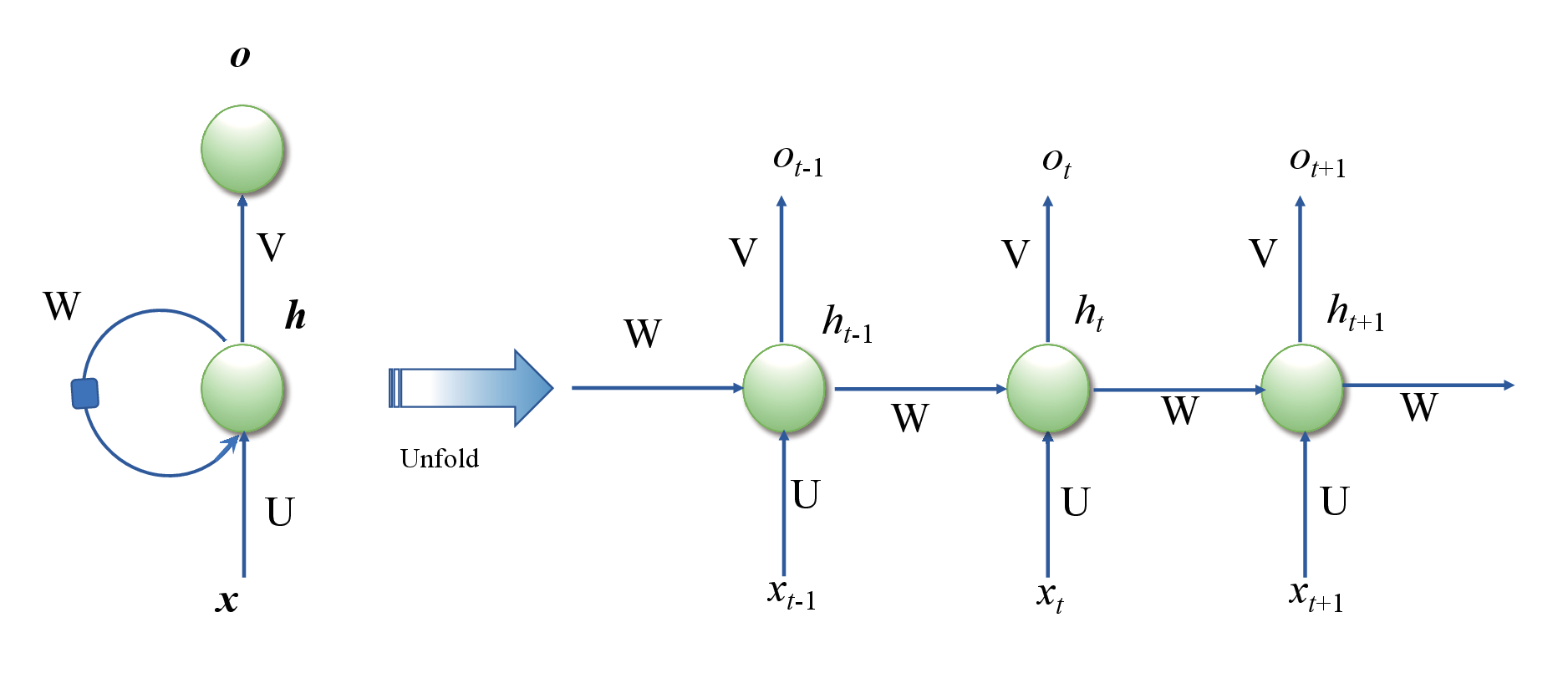}}
\caption{Unfolded basic RNN. (Refer to \cite{Deeplearning})}
\label{basicRNNarch}
\end{figure}

RNNs have  many variants, such as long short-term memory model~\cite{Hochreiter1997LSTM}, which can even be used to predict COVID-19~\cite{TYCBZheng2020}. Here we just discuss the basic RNNs.  Basic RNNs are  networks of neuron-like nodes organized into successive layers. Fig.~(\ref{basicRNNarch}) shows an unfolded basic RNN.

The mathematical expression of a basic RNN can be found in~\cite{Lipton2015RNN}
\begin{eqnarray}
\label{RNNfig1}
\mathbf{h}(t) &=&\sigma_h  ( U \mathbf{x}{(t)} + W \mathbf{h}{(t-1)} + \mathbf{b}_h) \nonumber\\
\mathbf{o}{(t)} &=&\sigma_o ( V \mathbf{h}{(t)} + \mathbf{b}_o).
\end{eqnarray}
This form belongs to Elman network~\cite{Elman1990} actually.


The output $\mathbf{o}{(t)} = \sigma_o ( V \mathbf{h}{(t)} + \mathbf{b}_o)$ is a single layer full connection network. Such kind of structures can be generated with Eq.~(\ref{CNNgen}) which has been presented at Subsection~(\ref{subsec:mlp}).  In the following, we will discuss the hidden layer output $\mathbf{h}(t)$.

Suppose that $\underbar {D}$ in Eq.~(\ref{CNNgen}) is a diagonal matrix,  and 
$\triangle u =[ \nabla_{T} ^2 + \partial ^2 /\partial z ^2]u$.  Now letting $\tau = z - v t $ (traveling wave case), we have
\begin{equation}
\label{RNNgen}
 -v \frac{\partial u}{\partial \tau} = D_{xy} \nabla_{T}^2 u + D_{z} \frac{\partial ^2}{\partial \tau ^2} u +f (u, \tau). 
\end{equation}
Here, the diffusion coefficient  $D_{x, y}$ means that $D$ only depends on variables $x, y$, and 
$D_z$ indicates that it only depends on variable $z$.  

When combining forward difference and a second order central difference for  $\tau$ variable,  we get the recurrence equation
\begin{eqnarray}
\label{RNNeq1}
 -v \left [\frac{ u_{\tau+1}- u_{\tau}}{k} \right ] &=& D_{xy} \nabla_{T}^2 u_{\tau} + \\
 &+& D_{z}\left [ \frac{u_{\tau+1}- 2u_{\tau}+ u_{\tau-1}}{ h ^2}  \right ] +f (u, \tau).  \nonumber
\end{eqnarray}
Note that, with the variable substitution (in the traveling wave case),  $\tau= z- v t$,  the difference $u(t+1)- u(t)$ will be expressed as $u_{\tau+1}- u_\tau$, and  $u(z+1)- u(z)$  is also expressed as $u_{\tau+1}- u_\tau$.  Moreover, the function $f(u, z, t)$ can be written in the form $f(u; \tau)$.

Eq.~(\ref{RNNeq1}) can be rewritten in the form~\cite{Hughes2019}:
\begin{equation}
\label{h_dinfnition}
\begin{bmatrix}
      u_{\tau+1}    \\
     u_{\tau}
\end{bmatrix} =
\begin{bmatrix}
     W_1 & W_2   \\
      W_3 & W_4 
\end{bmatrix} \begin{bmatrix}
      u_{\tau}    \\
     u_{\tau -1}
\end{bmatrix} + U  \begin{bmatrix}
     f (u, \tau)    \\
     0
\end{bmatrix}.
\end{equation}
where 
$$
 W_1 = M ( k h^2 D_{xy} \nabla _{T}^2 -2 kD_z - h^2 v), \,\,\,\,\, W_2 = - k M D_z,
$$
$W_3=1$, $W_4=0$, and $U= -k M {h^2}$. And $M=(v h^2+ kD_z)^{-1}$. Here  $k$ and $h$ are the time step value, and the spatial grid size value in FDM, respectively.

We define 
\begin{equation}
\mathbf{h}(\tau) = \begin{bmatrix}
   u_{\tau+1}    \\
     u_{\tau}
\end{bmatrix}.
\end{equation}
We then rewrite Eq.~(\ref{h_dinfnition})  in the form
\begin{equation}
\label{RNN-LNA}
\mathbf{h}(\tau) =   W \mathbf{h}{(\tau-1)} +  U \mathbf{f}{(\tau)}.  
\end{equation}
Especially, when we still consider information diffusion in a finite time, quasi-linear approximation can be applied to NPDEs. Otherwise, the above equation is the case of assuming the activation function $\sigma_h$ to be a linear function. It is worth noting  that NPDEs contain spacial and time variables,  which can also be applied to describe spatiotemporal dynamic systems~\cite{Wang2019TCYB}.

\section{NPDE Solver}
\label{NPDEslover}
In this section, we present algorithms for NPDEs. These discussions will be first set out from the perspective of supervised learning, and unsupervised learning. We will then suggest a new PDE-constrained optimization method for NPDEs.

Linear or semi-linear PPDEs are well studied in the literature, but only a few simple cases can obtain analytical  solutions, such as Fisher's equation~\cite{Roman2017}. An integral transformation may change a PDE to a simpler one, in particular, a separable PDE. This corresponds to diagonalizing an operator~\cite{Roman2017}.  In the book of ``{\it Nonlinear Reaction-Diffusion-Convection Equations}''~\cite{Roman2017},  an important example of integral transform is shown, which   is Fourier analysis.   Fourier analysis diagonalizes the heat equation using the eigenbasis of sinusoidal waves. In the case that the domain is finite or periodic, an infinite sum of solutions such as a Fourier series is appropriate, but an integral of solutions such as a Fourier integral is generally required for infinite domains. The solution for a point source for the heat equation  is an example of the utilization  of a Fourier integral.  For solving classical RDE  problems,  extensive studies have been conducted.  
Investigations of various initial and boundary conditions can be found in the literature,  for example, in~\cite{PaoCV1992nonlinear}\cite{cherniha2017nonlinear}.

However, the traditional PPDE solver requires that the system configuration function such as diffusion matrix should be given, while  for NPDEs we suppose that the diffusion matrix contains parameters which are learnable. Namely, the optimization algorithm should be adopted to develop the NPDE solver.  With physical interpretation of our model, we imagine that   the information wave passes through a layered medium, and the state is described with NPDEs. When solving NPDEs with numerical methods, we can use the nine-point stencil (in 2D case) finite-difference formula to find numerical solutions.   When input information is given, this can be regarded as the initial condition for NPDEs.  Because we do not know the medium parameter matrix $A(u, x, t),  B(u, x, t)$,  we can simply initialize  them with random values.  As for $C(u, x, t)$,  we can assume it as a nonlinear operator, which plays the role such as  activating operation and/or neurons pooling.
 Then we can compute the value of  $U$  at $t=T$,  and get  $U|_{t=T}$.  With these values, we can compute the loss function $L(\cdot )$. Consequently, we can exploit another PDE (refer to Eq. (\ref{equationTur1b}) in Appendices) to find the network parameter matrix $\theta$, it is equivalent to determining coefficient matrix $A(u, x, t), B(u, x, t) $.

 From these descriptions, it is evident that NPDEs are forward scenarios, while optimization algorithms are backward scenario.  
 In Appendices,  
 Eq.~(\ref{equationTur1}) describes these forward  and  backward scenarios mathematically. Once these configuration matrices are learned, NPDEs are transformed into traditional deterministic PDEs.  As almost all physical laws are described by PDEs, the physics systems are deterministic because they obey PDEs.  A real physical process  governed by the PDEs will have a deterministic solution. Therefore,  when  the initial and boundary conditions are obtained in advance,  solving analytically or numerically these PDEs means that we can predict the future of a physical system  all the times. 

In the following, we present several learning algorithms for finding numerical solutions of NPDEs.

\subsection{Supervised Learning for NPDEs}
\label{SL4NPDEs}
Those classification/regression problems usually belong to supervised learning cases.  We can write the loss function in the $L_2$ norm with the weight decay regularization.  Referring to \cite{Bishop2006}, we can formulate the loss function as follows:
\begin{equation}
\label{lossfunction1}
L( U( t=T), \theta) =\frac{1}{2}\left [  \|  U (t=T, \theta) - \mathbf{o}\| ^2  + \nu \| \mathbf{w}\|^2 \right ],
\end{equation}
where $ U (t=T)$ is the network output at the $t=T$,  $\mathbf{w}$ is the weight parameter, $\theta$ stands for a parameter family including hyper-parameters as well as weight parameters, and  $\mathbf{o}$ is the expected output.

The algorithm can be described as follows {\bf Algorithm \ref{Alg1}}.



\begin{algorithm}
\caption{Algorithm to solve NPDEs in supervised learning.}
\label{Alg1}
\KwData{initial and boundary condition $U(t=0, \mathbf{x}) $ and $U(t, \mathbf{x}|_0^1) $, respectively. Here $\mathbf{x}|_0^1$  means that the range of $\mathbf{x}$ is between 0 and 1.}

\KwResult{Randomly set parameter $\theta$, discretized elliptic operator value with step size $k$ and $h$.}

\While{Not Converge}{
Randomly set parameter $\theta$, discretized elliptic operator value with step size $k$ and $h$.

Numerically solve NPDE, obtain the solution $U(t=T)$.

Optimizing loss function Eq. (\ref{lossfunction1}) with either Adaptive moment estimation (ADAM) or L-BFGS algorithm to obtain parameter $\theta$.
}

\end{algorithm}


Under the framework of  the {\bf Algorithm \ref{Alg1}},  we can consider the spacial case, that is traditional DNN architecture  case. We know that he full connected MLP is equivalent to the full size kernel convolution in addition with the multi-channel operation. When applying FDM to NPDE, we just simple stack each element (building block) obtained by FEM (Eq. \ref{channel-1}), the MLP  architecture is generated. This can be considered as a trivial solution our proposed NPDEs. Experiments in our work 
\cite{BPIL2020} show the solid support for our proposed theory in this article.

Also, when applying FDM to NPDE with method described in subsection \ref{CNN-discribe}, and taking $k=1$, we can generate ResNet   architecture. And a deep CNN architecture can be generated by properly applying FEM  \cite{deng202174}. Experiments in \cite{deng202174} also show the solid support for our proposed theory.

More details about the learning strategies  and classical optimization algorithms, such as ADAM, L-BFGS, are presented in Appendices.


\subsection{Unsupervised Learning for NPDEs}
In~\cite{Synergetic2019}, it is discussed that a  system  can be described with nonlinear RDEs.   In two-model SLS shown in Fig.~(\ref{two-model-SYS}),  NPDEs can be used to model a generative network,  and its inverse model is the reductive network~\cite{Synergetic2019b}. This framework can be utilized as an unsupervised learning framework, since we can build an encoder-decoder framework with the two-model SLS. In this encoder-decoder framework, System A is assumed to be a decoder, while System B is the encoder.  In practice, we can apply the approximated synergetic learning algorithm in order to reduce the system's complexity~\cite{method-SLS} \cite{BPIL2020}.

\subsection{PDE-constrained Optimization}
The above discussion can be considered as a traditional nonlinear PDE solver. Here we suggest developing a new 
 PDE-constrained optimization method for  parameter learning in  NPDEs.
 
The idea is to adopt some metrics to measure the difference between the expected system output and real system output, while the system is constrained by NPDEs. 

 PDE-constrained optimization is a subset of mathematical optimization where at least one of the constraints may be expressed as a PDE~\cite{TrendsPDE2014}. 
 A standard  formulation of PDE-constrained optimization is given by \cite{FrontiesPDE2018}:
\begin{equation}
 \min _{y,u}\;{\frac {1}{2}}\|y-{\widehat {y}}\|_{L_{2}(\Omega )}^{2}+{\frac {\beta }{2}}\|u\|_{L_{2}(\Omega )}^{2},\quad {\text{s.t.}}\;{\mathcal {D}}y=u, 
\end{equation}
where $u$ is the control variable,  $\beta$ is the regularization parameter, and $ \|\cdot \|_{L_{2}(\Omega )}^{2}$   is the Euclidean norm. Closed-form solutions are generally unavailable for PDE-constrained optimization problems because of the complexity. Therefore, numerical methods~\cite{Antil2010} are necessary.

When we adopt the PDE-constrained optimization method, we need  model the PAI systems with a formal mathematical expression. In this work, we write the optimization equation as

\begin{eqnarray}
\label{PDE-constriant}
\min _{u, \theta}L(u, \theta)= \frac {1}{2} \|u -{\widehat {u}}\|^{2}+\frac {\beta }{2}\|\theta\|^{2},\quad \nonumber \\
{\text{s.t.}}\;  u_t(x, \theta, t ) - \mathcal{O}_L u(x, \theta, t) =0,
\end{eqnarray}
where $u$ is the system output, $ {\widehat {u}}$   is the expected system output, and $\theta$ represents the system configuration parameters (network parameters).  In the equation constraint term, $u_t(x, \theta, t ):= \partial {u}/{\partial t}$, and  $\mathcal{O}_L u$ is the same as Eq.~(\ref{neuralPDE1}). This loss function has the form of the first order Tikhonov  regularizer~\cite{Tikhonov1977}\cite{Guo2003reg}.

For supervised learning, ${\widehat {u}}$ will be  data labels.  While for the unsupervised learning, we can design  an autoencoder system, and $ {\widehat {u}}$ will be assumed as a given data representation function. In practice, there are two strategies that can be used to deal with this kind of optimization problems. That is, {\it Discretize-then-optimize} or {\it Optimize-then-discretize}. More details about this topic can be refereed to in~\cite{FrontiesPDE2018}.

\section{NPDEs Evolution To nmODE}

Let us briefly review the Brownian motion problem in statistical physics firstly.

\subsection{ Diffusion of a Brownian Particle in a Tilted Periodic Potential}

In statistical physics, Brownian motion is one of the stochastic process, this process is called diffusion.  The Brownian motion is extensively studied 
in the field of non-equilibrium statistical physics. While in the research field of  AI, inspired by non-equilibrium statistical physics, 
 Sohl-Dickstein  {\it et al} \cite{sohl2015deep} proposed the diffusion generative model in 2015.

 Here we mainly review thermal diffusion in a tilted (over-damped) periodic potential problem, which has 
a prominent role in various systems.        Thermal diffusion  is regarded as a simple and typical
problem in nonlinear-nonequilibrium physics.  The time evolution of the position $\mathbf{x}$ of the Brownian particle itself is  described using Langevin equation, which involves a random force field representing the effect of the thermal fluctuations.

The original Langevin equation is as follows \cite{Fokker-PlanckEq1996}:

\begin{equation}
 m{\frac {d\mathbf {v} }{dt}}=- \gamma \mathbf {v} +{\mathbf {R}}(t).
\label{Langevin}
\end{equation}

Where $m$ is the particle mass, $\gamma$  is the damping constant, and ${\mathbf {R}(t)}$  stands for a random force.
$\mathbf {v} = d\mathbf {x} /{d t}$ can be regarded as particle velocity.

When we consider  Brownian motion of particles   in a tilted periodic potential,  the
Langevin equation has the form:

\begin{equation}
\label{LangevinEq}
\frac{d\mathbf {v} }{dt} +\gamma \mathbf {v} + \frac{\partial V}{\partial x} =F_c + {\mathbf {R}(t)}. 
\end{equation}

Here $V(\mathbf {x}) = V(\mathbf {x} + L )$ is the periodic potential with period $L$ divided by the mass $m$
of the particle, $F_ c$  is the external force divided by $m$, and 
The random force ${\mathbf {R}(t)}$ describes white noise with zero mean (Wiener process).  The total potential is $U(\mathbf {x})= V(\mathbf {x})- F_c \mathbf {x}$.




When   a Brownian particle movement  subjected to the external periodic potential $V(x)$, consisting of attractive
well regions separated by potential barriers of height $V_0$ is considered,  a potential that with 
these characteristics is given by
\begin{equation}
\label{potentialV}
V(\mathbf {x}) = -\frac{V_0}{2}\cos(2\pi \mathbf {x}/ L).
\end{equation}

The Langevin equation can be reformulated as a Fokker--Planck equation (FPE),  which describes 
 the time evolution of the probability density function of the velocity of a particle under the influence of drag forces and random forces. In one-dimensional case the FPE has the form\cite{Fokker-PlanckEq1996}:

\begin{equation}
\label{FP-equation}
\frac{\partial p({v} , t)}{\partial t} =\left[ - \frac{\partial  {v}}{\partial  {x}} +\frac{\partial }{\partial {v}} 
\left(\gamma  {v} + \frac{\partial U({x})}{\partial  {x}} + \gamma \Theta \frac{\partial }{\partial  {v}} \right )p( {v} , t)\right].
\end{equation}
Where $p({v} , t)$ is used to express the probability density function,  the thermal velocity $v_{th} = \sqrt{k_B T /m}$ .   $k_B$ is the Boltzmann constant, $\Theta= v_{th}^2$ and $T$ is temperature.

The Fokker-Planck Equation is a generalization of the diffusion equation,  the diffusion equation is a parabolic PDE.





\subsection{nmODE}

The neural ordinary differential equation (neuralODE) was developed by Chen {\it et al}~\cite{ChenRBD2018ode} in 2018,  it is used to describe the dynamics of a neural network. 

A general neuralODE can be described mathematically as
\begin{equation}
\label{2eq:neuralODE}
\frac{d u}{dt } = F (u,  x)
\end{equation}
for $t \geq 0$, where $u \in R^{n}$ denotes the state of the network, $x \in R^{m}$ represents external input variable and we take it for data inputting, the function $F$ is continuous and satisfies some Lipschitz condition. 
 Given an initial condition $u(t=0)$, the  equation (\ref{2eq:neuralODE}) can be solved with numerical simulation if analytical form solution cannot be obtained. 
 
The concept of neural memory ordinary differential equation (nmODE) was introduced by Zhang Yi  \cite{zhangyi2023nmODE} in 2023,  which is  a type of ordinary differential equation that represents a continuous dynamic system.  The nmODE  is expressed mathematically  as follows:

\begin{equation}
\label{2nmODE}
\frac{d u} {d t} = - u + \sin^{2} \left[ u + W x + b \right]
\end{equation}
for $t \geq 0$, where $u \in R^{n}$, $x \in R^{m}$, $b \in R^{n}$, $W=\left(w_{ij}\right)_{n \times m} \in R^{n \times m}$. 

In  neural networks  viewpoint, $x$ represents external input to the network,  $u(t)$  denotes the state of the network at time $t$, and $u(t=0)$ is the initial value.   The external input variable $x$  stands for  an input neuron, while any neuron used to represent the state $u$ is called a memory neuron. And  the $\sin^2(\cdot)$ function represents  the activation function.

When the system reaches an equilibrium point, $ d u/{d t} = 0$. Then  the equilibrium equation of the nmODE can be obtained,  \begin{equation}
\label{2pImEqn}
  u = \sin^{2} \left[ u + W x + b \right].
\end{equation}
This is an implicit mapping equation, which defines a nonlinear mapping $F$ from input $x$ to output $u$ as:
\begin{displaymath}
F: x \rightarrow u.
\end{displaymath}

When taking Euler's method to Eq. (\ref{2nmODE}), we have
\begin{equation}
\frac{u(t+k)- u(t)} {k} = - u(t) + \sin^{2} \left[ u(t) + W x + b \right],
\end{equation}
where   step parameter $k$ is a hyperparameter in neural network, $0 < k <1$.  The   discrete version of the nmODE is obtained when  using $u_{l+1}$ denotes $u(t+k)$, $u_{l}$ denotes $u(t)$, and replaces $k$ with $\epsilon$, we have

\begin{equation}
\label{epsilon-net}
u_{l+1} =(1 - \epsilon) \cdot u_l + \epsilon \cdot \sin^{2} \left[ u_l + W x + b \right],
\end{equation}
 at here $l$ stands for  layer number of neural network.  Eq. (\ref{epsilon-net})  is called as $\epsilon$-network.

\subsection{NPDE to nmODE}

NPDE can be applied to generate neural architectures, in a special situation it can evolve to nmODE also. Following we will present how NPDE evolve to nmODE.

In mathematics, FPE is  called  the Kolmogorov forward equation, while  in the inverse procedure (time inverse)  the Kolmogorov backward equation  is useful \cite{Fokker-PlanckEq1996}.

Inspired by the diffusion process and condensation process in statistical physics,  Guo proposed the Maxwell's demon technique to optimize two-model SLS \cite{Guo2023icmlc} in 2023.  In that work, one NPDE, (similar with Kolmogorov forward equation), is used to describe generative model. And another NPDE, (similar with Kolmogorov backward  equation),  is used to describe reductive model.

Here we discuss NPDE to nmODE evolution problem. Suppose information particles diffuse in a tilted periodic potential, the diffusion probability model is described by a FPE-like NPDE. When we consider dynamics of information particles, the neural ODE (similar to Langevin equation) is adopted.

In our neural model, we suppose the periodic potential $V(z)=\frac{1}{4}\sin(2z)$,  the  potential is
tilted by a homogeneous, static force $F_c=1/2$, and random force $R(t)=0$.   Similar with  Eq. (\ref{LangevinEq}), we can obtain

\begin{eqnarray}
\label{neuralLangevinEq}
\frac{d v }{d t}&=& -\gamma {v} + 1/2 - \frac{\partial V(z) }{\partial z}   \nonumber \\
                           &=& -\gamma {v} + \frac{1}{2}(1-\cos(2 z)) \nonumber \\
                           &=& -\gamma {v} +\sin^2(z).
\end{eqnarray}
Now turning  to neural model, we let $z=v+ Wx +b$ is the input of  an activation function $\sin^2(\cdot)$,  the Eq. (\ref{neuralLangevinEq}) is exact same with Eq. (\ref{2nmODE}) (nmODE) when $\gamma =1$.

Here we can see that general NPDEs evolve to  nmODE in the special scenario  of information particles diffusion  in a tilted periodic potential. Also, we provide an alternative explanation  for nmODE  based on the  non-equilibrium statistical physics.


Besides NPDE can describe the spatiotemporal dynamic neural systems,  it can be applied to neural architecture evolution problem.
In subsection \ref{RNNsubsection},  we present the NPDE (Eq. (\ref{RNNgen})), which has the form of PPDE.  At quasi-linear approximation with the linear activation function, NPDE generates a RNN type neural architecture as shown in Eq. (\ref{RNN-LNA}).

When we take back to the nonlinear activation function and still consider that information particles diffuse in a limited speed, the  Eq. (\ref{RNN-LNA}) can be returned back to RNN network (Eq. (\ref{RNNfig1})). If assuming $\sigma _h (\cdot) =\sin ^2 (\cdot) $ and
$\mathbf {o}(t) =  \mathbf {h}(t) $ in Eq. (\ref{RNNfig1}),  we can easily find that Eq. (\ref {2pImEqn}) has the similar form.  This illustrates that neural architecture  nmODE described belongs to RNN type.





\section{Discussion}
\label{sec:discussion}
In this section, we would like to conduct important discussions and provide some insights into NPDEs.

By utilizing  FDM along with FEM, we discrete NPDEs to generate three popular building blocks for DNN architectures. The advantages of neuronizing PDEs have two folds.
\begin{enumerate}
\item [1)] A theoretical foundation for applying  AI to solve mathematical physics PDEs is built up; this offers great potentials to be applied extensively e.g. in the fields of physics and chemistry.   Some tricks in deep learning, such as the paddle method including extended, wrap, mirror, and crop,  can be used to deal with the boundary condition problem. 
\item[2)]  NPDEs can be exploited to describe the AI systems;  this may lead to interpretable AI based on statistical mechanics. NPDEs can be regarded as the  theoretical foundation for  PAI systems, which further provides theoretical guidance for designing AI analog computing  devices. 
\end{enumerate}
Interpretations about these two folds are further discussed as follows.
\subsection{AI for PDEs}
Though it has been studied for decades to use AI to solve PDEs, it remains unclear why AI can be applied to the PDEs described problems successfully. Our work demonstrates that when adopting FDM (along with FEM) to solve PDEs numerically, it is basically equivalent to using DNN to do so. In this sense, our work provides a theoretical foundation for applying AI to solve PDEs.

Recently, great success has been attained in applying AI to solve PDEs. For example,  Han {\em et al}. used deep learning to solve high-dimensional PDEs~\cite{WeinanPNAS2018}, including solving Hamilton-Jacobi-Bellman equation for control. They also applied DNN to solve  many-electron schr\"{o}dinger equation~\cite{HanJCP2019}. More recently,  Pfau {\em et al}. have developed a new neural network architecture, named FermiNet,   to model the quantum state of large collections of electrons~\cite{PhysRevResearch2020}. They believe that 
FermiNet presents the first demonstration of deep learning for computing the energy of atoms and molecules from the first principles. They also showed how deep learning can help solve the fundamental equations of quantum mechanics for real-world systems.

\subsection{NPDEs for AI}
In this work, we propose general NPDEs as shown in Eq.~(\ref{neuralPDE}).  It can be considered as a fundamental equation for PAI systems, for example,  synergetic learning systems described with nonlinear RDEs~\cite{Synergetic2019}.

\subsubsection{Interpretable AI}
Using NPDEs to describe AI systems implies explainable AI systems.  Taking an SLS (shown in Fig. (\ref{two-model-SYS})) as one typical example, we describe the  PAI systems in the following.

SLS can be regarded as a simple PAI system model,  in which the information propagates, and diffusion and reaction occur.  An information wave is different from an electromagnetic wave, which is the carrier of information, but the elliptic and parabolic equations in  mathematical physics equations can be used to describe the process of information transmission. Methods to solve such equations depend on the complexity of the problem. At present, most studies on nonlinear PDEs in mathematics take the FDM (along with  FEM) to obtain numerical solutions. In \cite{Guo1990AOS} \cite{Guo1990} \cite{Guo1999Dynamics},  we applied the heat diffusion equation to study the propagation of light beams in nonlinear media and the dynamics of interference filters. The diffusion equation is a semi-linear PPDE and can also be used to study dispersive optical tomography~\cite{Niu2008Improving32}.
 
\subsubsection{Analog Computing Device Design}
To realize a PAI system,  hardware computing components are necessary to assemble it. Currently, photonics computing presents one potential filed~\cite{light2021}.  More introduction about analog computing device design topic has been discussed in Appendices.

\subsubsection{Materials Design} 
  NPDEs can be applied to materials design for PAI systems, which is named ``inverse design''.  
New materials  design  is to find those unknown coefficients $A(x, t)$, $B(x, t)$ and $ C(x, t)$ in Elliptic operator (Eq.~(\ref{ellopticopra})).

When obtaining numerical values of those unknown functions in each time slice, we model them with a functional expression. The obtained functions can be used for materials design.  This is an interesting topic worthy of further study. We refer the interested readers to Sanchez-Lengeling \& Aspuru-Guzik's  review paper ``{\em Inverse molecular design using machine learning: Generative models for matter engineering}''
\cite{Sanchez2018}, which describes generative models for matter engineering. 

\subsection{Insight into NPDEs}
We give more discussions about NPDEs in this subsection.

 \subsubsection{Difference with PDE Nets}
 As discussed in~\cite{Guo2020archEvolve},  RVFL  and MLP with direct link can evolve to an NPDE; it is also shown that ResNet can evolve to  an NPDE with  Elliptic operator.   Previous  research works, such as  ODE and PDE net~\cite{ChenRBD2018ode}\cite{LongLD2019PDEnet} \cite{LuZLD2018icml}, are transformations from discrete   to continuous variables. 
 By taking FDM along with FEM,  NPDE is actually a transformation from   continuous to discrete variables to make the problem  computable.  Moreover, discretization is  the process of transforming continuous functions, models, variables, and equations into discrete counterparts. This procedure can be regarded as quantized problem as well.  In the numerical solution of PDEs,  there are two hyper-parameters $k$ and $h$ in FDM,
 which control the depth and width of the neural network structure, respectively. Also, in the  literature~\cite{numericalRDE}, we know that  $k$ and $h$ control the error of PDEs' solution.  
 
  Discretizing (quantizing)  PDEs can  construct extremely deep and extremely wide  neural architectures. This depends on the parameters  $k$ and $h$ selection when FEM is applied. While PDEs with continuous variables can be applied to describe infinite deep and infinite wide  neural architectures instead of infinite wide only neural networks~\cite{Neal1996Infinite}.    In Eq.~(4) of ResNet paper~\cite{eccvHeZRS2016ResNet2}, the summation term can be transformed  to an integral form when the network's depth approaches infinite.  Conversely, under the sparse approximation, we can write an integral as  summation~\cite{Guo2002}.  The sparse approximation means that we should assign relative large  $k$ and $h$ parameters under a given  finite volume of medium. 

\subsubsection{Realizable  PAI  Systems}
In order to realize PAI systems, solid theoretical foundation should be first set out.  It is notable that unlike  ODEs, essentially all PDEs that have been widely studied come quite directly from physics. For example, the diffusion equation arises in physics from the evolution of temperature or of gas density;  the wave equation represents the propagation of linear waves, for instance,  along a compressible spring or along fluids; RDEs come from studies of dissipative structure, representing far from thermal equilibrium state in thermodynamics.  In this work, we introduce NPDEs to describe the PAI systems. We highlight several remarks as follows.

\begin{itemize}
\item  Inspired by Blackbody radiation theory, radiated  spectra  is continuous from macro view, and is quantification from micro view. Therefore, while the mathematical method is FDM, its physical basis  is quantization (discretization).
\item When  realizable PAI  systems are designed based on the SLS architecture, it is mandatory that   at least one sub-system should be a real part, instead of two virtual  sub-systems. The real part sub-system  consists of some components which can implement optical computation, e.g., with materials of  nonlinear optical medium.
\item The approximated method for our developed nonlinear to quasi-linear PDEs  makes NPDEs solvable. Quasi-linear approximation is reasonable because it is   well founded physically. According to the special relativity theory, there is no possibility of superluminal velocity. Consequently, information propagation in a PAI system has a  limited speed that is slower than light.
\item Strictly speaking, there are no absolute  simultaneous  computing convolution and activation/pooling operation in a PAI system as limited by the physics law.  Hence, in a PAI system,  asynchronous computing is real case, and time delay should be taken into account when solving NPDEs. Both asynchronous and time delay  should be considered when we design a realizable PAI system.
\item In the physical world, causality is related to time, because ``causes'' cannot happen after the ``consequences''. If we believe that the  arrow of time is unidirectional,   the NPDEs with time variable will allow us to  predict the
future behavior of a PAI system.  It also  allows us to gain insight in a PAI system, and develop an interpretable DNN for causal inference.  
\end{itemize} 

\subsubsection{NPDE Solution} First principle in  quantum chemistry  is {\it ab initio  } method,  which solves Schr\"{o}dinger equation. In AI systems, we believe that first principle is PLA, which solves NPDEs.  

To solve  NPDEs numerically, we should pay attentions to the following questions:
\begin{itemize}
\item How can we  set up initial conditions and boundary conditions for various real world applications? 
\item How can we  choose step parameters $k$ and $h$, i.e. how to  trade-off between the computation accuracy and complexity?
\item By applying FEM to NPDEs, how can we establish the relationship between  Capsulate networks and elements obtained by FEM?
\item Compared with neural  architecture search methods, how can we generate an optimal neural  architecture for the given condition (data set) with Turing's equations?  
\end{itemize}

Currently, we can use numerical methods to find NPDEs' solution. However, like Navier-Stokes equations,  which are fundamental equations of fluid mechanics, it is  very challenging mathematically to find analytical solutions. It remains an open problem on how to solve NPDEs analytically.

\section{Summary}

Based on the  first principle, we take the free energy as the action, and  derive the RDE to describe SLS. In this work, we propose the general  NPDEs  to describe PAI systems. By utilizing  FDM along with FEM, we also generate three most popular DNN building blocks by discretizing PDEs.  

In addition to the discussion of  generating neural architectures with NPDEs, we study the AI system optimization algorithm.  The learning strategies, such as ADAM, L-BFGS,  pseudoinverse learning  algorithms, and PDE- constrained optimization methods, are  presented in  Appendices. Based on the physics law, we also propose the quasi-linear approximation to solve nonlinear PDEs.

Most significantly, we present a clear physical view of interpretable DNNs. Namely,  an AI system is interpreted as information diffusion and reaction processing.  When information is considered as wave propagation in an AI system,  analog optical computing device is suitable to construct a PAI system.

While classical PDEs describe the physics systems,  NPDEs can be used to describe the PAI systems.  By utilizing  FDM along with FEM, we  bridge the gap  between mathematical physics  PDEs and modern DNN architectures.   In other words, DNNs can be obtained by properly applying  discretization to NPDEs.  In this viewpoint, we believe that current DNNs are special cases of  NPDEs. Furthermore, we conjecture that  NPDEs could be the fundamental equations for  PAI systems, which may offer a theoretical basis for  analog computing device design.

\begin{acknowledgement}
The research work described in this chapter was fully supported  by the National Key Research and Development Program of China (No. 2018AAA0100203) and National Science Foundation of China (No. 61876155). 
\end{acknowledgement}

\section{Appendices}
%
%
\subsection{Appendix 1: Generating Restricted Boltzmann Machine}
Here we present that Restricted Boltzmann Machine (RBM) can be generated with stochastic partial differential equations.

With quasi-linear approximation, we obtain the neural partial differential  equation 
\begin{eqnarray}
\label{neuralPDE1a}
 \frac{\partial \Psi }{\partial t}&=&\mathcal{O}_L \Psi, \qquad \mbox{where} \\
\mathcal{O}_L \Psi  &=&\nabla \cdot (A ( \Psi )\nabla \Psi)+ B( \Psi ) \nabla \Psi + C(\Psi) \nonumber.
\end{eqnarray}

When using  the FDM (along with FEM) to discrete RDEs,   we know that if  using a forward difference at time $ t_{n}$  and a second-order central difference for the space derivative at position $x_{j}$,  we get the recurrence equation:
\begin{equation}
\label{heatEq}
\frac {u_{j}^{n+1}-u_{j}^{n}}{k}={\frac {A_{j+1}u_{j+1}^{n}-2A_{j}u_{j}^{n}+A_{j-1}u_{j-1}^{n}}{h^{2}}}.
\end{equation}
This is an explicit method for solving the one-dimensional inhomogeneous heat equation with the diffusion matrix $A(x, t)$.

We can obtain $ u_j^{n+1}$  from the other values in this way:
\begin{eqnarray}
\label{heatEq2}
 u_{j}^{n+1}&=&(1-2rA_{j})u_{j}^{n}+rA_{j-1}u_{j-1}^{n}+rA_{j+1}u_{j+1}^{n},\nonumber \\
                      &=& \sum _j W_j^n u_{j}^{n}.
\end{eqnarray}
where $r=k/h^2$. 

 If we use the backward difference at time $ t_{n+1}$  and a second-order central difference for the space derivative at position $x_{j}$  we get the recurrence equation:
\begin{equation}
\label{heatEq3}
\frac{u_{j}^{n+1} - u_{j}^{n}}{k} =\frac{A_{j+1}u_{j+1}^{n+1} - 2A_{j}u_{j}^{n+1} + A_{j-1}u_{j-1}^{n+1}}{h^2}. 
\end{equation}
This is an implicit method for solving the one-dimensional heat equation.

We can obtain $ u_j^{n+1}$  by solving a system of linear equations:
\begin{eqnarray}
 \label{heatEq4}
u_{j}^{n}&=& (1+2rA_{j})u_j^{n+1} - rA_{j-1}u_{j-1}^{n+1} - rA_{j+1}u_{j+1}^{n+1} \nonumber \\
                &=& \sum _j W_j^{n+1} u_{j}^{n+1}.
\end{eqnarray}

In the explicit method (forward Euler),  we regard $u_{j}^{n+1}$ in Eq.~(\ref{heatEq2})  as a hidden neuron, while in  the  implicit method (backward  Euler),  we take $u_{j}^{n}$ in Eq.~(\ref{heatEq4}) as a visible neuron.  It is noted that the traditional heat  equation  only contains Laplace operator.  When we take a general Elliptic operator on considering $A(x, t)$  and $c(x, t)$ as learnable coefficients,  $A(x, t)$  with discrete Laplace operator  stencil stands for the connecting weight. Moreover, one part $c(x, t)$ stands for the bias vector, and the other part is used as the activation function, if we split $c(x, t)$ into two parts.   Combining Eq.~(\ref{neuralPDE1}), Eq.~(\ref{heatEq2}) and Eq.~(\ref{heatEq3}), we can get the energy configuration of the RBM. Additionally, the free energy can be used as a loss function to perform the energy-based learning~\cite{Lecun2006A13}.

\subsection{Appendix 2: Analog Computing Device Design} 

In this section,  we   simply introduce optical analogy computing devices.

It is well-known that wave-particle duality has been demonstrated for photons (light), electrons and other microscopic objects (de Broglie wave). 
Here we make an hypothesis that information has wave-particle duality, since photons  have long been considered as the carriers of information in optical computing. Compared with electron computing, we believe  that optical analogy computing may display the advantages of  analog computer verse to  optical digital computer. 
Our previous works include Gaussian beam propagation in a nonlinear medium~\cite{Guo1990}\cite{Guo1990AOS}, and  properties of optical interference filter device~\cite{Guo1993HeatDyna}.

The advantage of using optical computing device is that 
with  photons as information carriers,  information propagates at the speed of light.  Also, when we take the passive transmission of optical waves, it can aid ultra-low power consumption. It is suggested in~\cite{Guo1987}\cite{Guo1999Dynamics} that optical interference filters can be considered as the candidates of the analog computing device. For example, ZnSe film filter can be used as optical bistability device~\cite{Sun1987}. But with a proper design, it can be used as a gate device to  play the role of sigmoid type activation function~\cite{Wang1987}\cite{Guo2001OMLNN}.

 We justify why  we would like to realize a PAI system with an optical  device.  From Fourier optics principles we know that when an optical system consists of an input plane, an output plane, and a set of components, the output image is related to the input image by convolving the input image with the optical impulse response $R$.  If we design an optical device with a layered micro-structure,  the input image $u_0(x, y)$ at the input plane (input layer) is
$$
u_0(x, y)= U(x, y, z){\big |}_{{z=0}}.
$$

The output plane (layer) is at $z = d$. The output image $u_d(x, y)$ is 
$$
u_d(x, y)=U(x, y, z){\big |}_{{z=d}}.
$$

The 2D convolution of input function against the impulse response function is
$$
u_d(x, y)~=~R(x, y) \ast u_0(x, y),
$$
i.e.,
$$
u_d(x, y)=\int _{{-\infty }}^{{\infty }}\int _{{-\infty }}^{{\infty }}R(x-x',y-y')~u_0(x, y)~dx'dy'.
$$
That is,  when we assume the optical impulse response $R$ to be a convolutional kernel, an optical system naturally plays a convolution operation.   In an imaging system, the optical impulse response $R$ plays a role of point spread function (PSF). Usually the PSF is given, but for blind de-convolution problem, PSF is unknown; this will become the convolutional kernel in CNN.

As Wu \& Dai said, {\em ``a hybrid optical–electronic framework that takes advantage of both photonic and electronic processors could revolutionize AI hardware in the near future'' }\cite{light2021}. 
 A detailed discussion about this topic is beyond the scope of this chapter.  Further research focus may be made on the design of NPDEs based nonlinear integrated photonic computing architectures~\cite{Xu2021ONN} \cite{Feldmann2021}.

\subsection{Appendix 3: Learning Strategies} 
\label{appendixA-LS}

\subsubsection{Learning Optimization}

Suppose the information $U$ along $z$ direction diffuse.  At the time $t_1$,  the system output is $U(z(t_1))$.   
We define the loss function $L( )=L( U(z(\tau_1))) $ and suppose that it is differential.  The optimal parameter $\mathbf{\theta} $ at given data set is 
 \begin{equation}
\mathbf {\theta} = \mathop{\arg\min} _{\theta} L (U(z(\tau_1)), \theta).
\label{SLS_principle}
\end{equation}

We can use  the gradient-based optimization (first order methods) methods, such as plain gradient, steepest descent, conjugate gradient, Rprop,
and stochastic gradient.  
When used to minimize the loos function, a standard (or ``batch'') gradient descent method would perform the following iterations:
\begin{equation}
\theta_{t+1} = \theta _{t} -\eta \nabla L(\theta),
\end{equation}
where $\eta$  is a step size ( learning rate).  From~\cite{Bishop2006}, we have
\begin{equation}
\frac{d \mathbf{\theta}}{d t} = -\eta \nabla_{\theta} L(\theta).
\end{equation}
This is an ordinary differential equation for network  parameter $\theta$.

There are several optimization methods, for example, genetic algorithm~\cite{TYCBSun2020}, but the most used method is the gradient descent based method. In the following, we will review the gradient descent algorithms.

\subsubsection{Gradient Descent Algorithms}

For the first order gradient descent algorithm, one  can adopt  Adam optimizer~\cite{kingma2014adam}. However, we would like to utilize   the second order optimization algorithm, for example,  Newton's method~\cite{boyd_vandenberghe_2004}:
\begin{equation}
\label{secondorderH}
\theta ^{(t+1)}=  \theta^{(t)} - \eta  (\mathbf {H}_ \theta ^{(t)})^{-1} \nabla_ \theta L^{(t)},
\end{equation}
where $\mathbf {H}_ \theta$ is a Hessian,
$$
\mathbf {H} _{i,j}={\frac {\partial ^{2}L}{\partial \theta_{i}\partial \theta_{j}}},\,\,\,\,\,\, \mathbf {H} = \nabla_ \theta ^2 L.
$$

In case of  reducing the complexity of the given problem, certain approximation for solving neural PDEs can be adopted.

{\bf   Pseudoinverse Approximation of Hessian}

In Newton's method, the Hessian is needed to be computed. But it is very computational intensive.
$\mathbf {H} _{i, j}={\frac {\partial ^{2}L}{\partial \theta_{i}\partial \theta_{j}}}$ ,  for a complex neural network, the parameter number is very large, hence the Hessian is huge.  Therefore we need to take some approximations to reduce the complexity.  Here we try to find some other methods, not just quasi-Newton method, to investigate the PDE network solver theoretically.

As we know,  if $\mathbf {H}$ is invertible, its pseudoinverse is its inverse. That is, $\mathbf {H}^{+}=\mathbf {H}^{-1}$\cite{numericalAN}.  Now we use the pseudoinverse instead of inverse in Eq.~(\ref{secondorderH}).
\begin{equation}
\mathbf {H}^{-1} = \mathbf {H}^{+} =(\mathbf {H}^{*}\mathbf {H})^{-1}\mathbf {H}^{*}.
\end{equation}
where $\mathbf {H}^{*}$ denotes Hermitian transpose (conjugate transpose). Since  $\mathbf {H}$ is a real and    symmetric matrix, it has $\mathbf {H}^{*}=  \mathbf {H}^{T}= \mathbf {H}$.   Nevertheless, we would like to keep the pseudoinverse form. Now Eq. (\ref{secondorderH}) becomes 
\begin{equation}
\label{pseudoinverseH}
\theta ^{(t+1)}=  \theta^{(t)} - \eta  (\mathbf {H}_ \theta^{*}\mathbf {H}_ \theta)^{-1}\mathbf {H}_ \theta^{*} \nabla_ \theta L^{(t)}.
\end{equation} 
 
 When we assume the matrix $\mathbf{A}(\theta) = (\mathbf {H}_ \theta^{*}\mathbf {H}_ \theta)^{-1}$,  Eq.~(\ref{pseudoinverseH}) can be written in the form:
 \begin{equation}
\label{pseudoinverseturing}
\frac{\partial \mathbf{\theta}}{\partial t}=   - \eta  \mathbf {A}(\theta)  \nabla_ \theta ^{2} L^{(t)} \nabla_ \theta L^{(t)}.
\end{equation} 
This PDE is used for optimization, which we call as the pseudoinverse learning method in finding network's parameters.

 {\bf  Symmetry Break}
 
 We rewrite Eq.~(\ref{neuralPDE}) 
 \begin{equation}
\label{neuralPDE2}
 \frac{\partial \Psi }{\partial t}=\mathcal{F}\left [ x, t,  \nabla, \nabla ^2, A ( \Psi ), B( \Psi ),  C (\Psi), \Psi \right ]
 \end{equation}
 and Eq.~(\ref{pseudoinverseturing}) as follows:
 \begin{eqnarray}
 \label{equationTur}
 \frac{\partial \mathbf {u}}{\partial \tau} &=& - A (\mathbf{u}) \nabla _{x}^2 \mathbf {u}(\tau) + (\nabla _{x} A(\mathbf{u}))(\nabla _{x}\mathbf {u}(\tau)) \nonumber\\
 &+& B(\mathbf{u}) \nabla _{x}\mathbf {u}(\tau) + C  \mathbf {u}(\tau), \nonumber \\
 \frac{\partial \mathbf{\theta}}{\partial t} &=&   - \eta  \mathbf {A}(\theta)   \nabla_ \theta ^{2} L^{(t)} \nabla_ \theta L^{(t)}.
\end{eqnarray}

We merge $\nabla _{x} A(\mathbf{u})$  and $ B(\mathbf{u}) $ into a new parameter $B_n(\mathbf{u})= \nabla _{x} A(\mathbf{u})+B(\mathbf{u})$, then we have:
\begin{subequations}
 \label{equationTur1}
 \begin{eqnarray}
 \frac{\partial \mathbf {u}}{\partial \tau} &=& - A (\mathbf{u}) \nabla _{x}^2 \mathbf {u}(\tau)   + B_n(\mathbf{u}) \nabla _{x}\mathbf {u}(\tau) + C  (\mathbf {u}(\tau)), \label{equationTur1a}\\
 \frac{\partial \mathbf{\theta}}{\partial t} &=&   - \eta  \mathbf {A}  \nabla_ \theta ^{2} L^{(t)} \nabla_ \theta L^{(t)}\label{equationTur1b}.
\end{eqnarray}
\end{subequations}

Unlike Turing's equations,  two PDEs in Eq.~(\ref{equationTur1}) are not symmetric, and they also have different meanings.  Variable $\mathbf {u}$ is in  the data space, while variable $\mathbf{\theta}$ is in the parameter space.
$\tau$ in the first equation stands for the depth direction in a virtual medium. In the second equation, $t$ means the iterative step in parameter optimization processing.

 {\bf  Gauss -- Newton Algorithm}

 In practical implementations, to avoid  computing Hessian directly,  Quasi-Newton methods are utilized. Quasi-Newton methods are  used to either find zeros or local maxima and minima of functions, as an alternative to Newton's method. They can be used if the Jacobian or Hessian is unavailable or too expensive to compute at every iteration. The ``full'' Newton's method requires the Jacobian in order to search for zeros, or the Hessian for finding extrema.

 The Gauss–Newton algorithm is used to solve non-linear least squares problems. It is a modification of Newton's method for finding a minimum of a function. Unlike Newton's method, the Gauss–Newton algorithm can only be used to minimize a sum of squared function values. It has the advantages that the second derivatives (challenging to be computed), are not required.
 
To minimize the sum of squares
 $$
 L({\boldsymbol {\theta }})=\sum _{i=1}^{m}r_{i}^{2}({\boldsymbol {\theta }}),
 $$
we  should  
start with an initial guess $\boldsymbol \theta^{(0)}$ for the minimum. The method then proceeds by the iterations,
$$
{\displaystyle {\boldsymbol {\theta }}^{(s+1)}={\boldsymbol {\theta }}^{(s)}-\left(\mathbf {J_{r}} ^{\mathsf {T}}\mathbf {J_{r}} \right)^{-1}\mathbf {J_{r}} ^{\mathsf {T}}\mathbf {r} \left({\boldsymbol {\theta}}^{(s)}\right),}
$$
where  $\mathbf {J_{r}}$ is  the  Jacobian matrix.

Note that $\left(\mathbf {J_{f}} ^{\mathsf {T}}\mathbf {J_{f}} \right)^{-1}\mathbf {J_{f}} ^{\mathsf {T}}$ is the left pseudoinverse of $\mathbf {J_{f}}$. This form looks the same as  Eq.~(\ref{pseudoinverseH}). In fact, it is one special case of the Newton's method.

In the family of quasi-Newton methods, the  
limited-memory Broyden- Fletcher - Goldfarb -Shanno algorithm (BFGS) (L-BFGS or LM-BFGS) is also  an optimization algorithm~\cite{Liu1989BGFS}\cite{nipsChenWZ2014}. 

Now we have two equations: one is for forward propagation, and the other is for back propagation. In addition to the L-BFGS algorithm, some other algorithms  can also  be used for optimization.

\bibliographystyle{spmpsci}
\bibliography{slsref2}

\begin{thebibliography}{10}
\providecommand{\url}[1]{{#1}}
\providecommand{\urlprefix}{URL }
\expandafter\ifx\csname urlstyle\endcsname\relax
  \providecommand{\doi}[1]{DOI~\discretionary{}{}{}#1}\else
  \providecommand{\doi}{DOI~\discretionary{}{}{}\begingroup
  \urlstyle{rm}\Url}\fi

\bibitem{FisherEQ1995}
Adomian, G.: Fisher-kolmogorov equation.
\newblock Applied Mathematics Letters \textbf{8}(2), 51 -- 52 (1995).
\newblock \doi{https://doi.org/10.1016/0893-9659(95)00010-N}.
\newblock
  \urlprefix\url{http://www.sciencedirect.com/science/article/pii/089396599500010N}

\bibitem{Antil2010}
Antil, H., Heinkenschloss, M., Hoppe, R.H.W., Sorensen, D.C.: Domain
  decomposition and model reduction for the numerical solution of {PDE}
  constrained optimization problems with localized optimization variables.
\newblock Computing and Visualization in Science \textbf{13}(6), 249--264
  (2010).
\newblock \doi{10.1007/s00791-010-0142-4}.
\newblock \urlprefix\url{https://doi.org/10.1007/s00791-010-0142-4}

\bibitem{FrontiesPDE2018}
Antil, H., Kouri, D.P., Lacasse, M., Ridzal, D. (eds.): Frontiers in
  {PDE}-Constrained Optimization, \emph{The IMA Volumes in Mathematics and its
  Applications}, vol. 163, 1 edn.
\newblock Springer-Verlag, New York (2018).
\newblock \doi{110.1007/978-1-4939-8636-1}.
\newblock \urlprefix\url{https://www.springer.com/gp/book/9781493986354}

\bibitem{BASSET1903}
BASSET, A.B.: The principle of least action.
\newblock Nature \textbf{67}(1737), 343--344 (1903).
\newblock \doi{10.1038/067343d0}.
\newblock \urlprefix\url{https://doi.org/10.1038/067343d0}

\bibitem{Bigot1987}
Bigot, J.Y., B., G.J.: Switching dynamics of thermally induced optical
  bistability: Theoretical analysis.
\newblock {Phys. Rev. (A)} \textbf{35}(9), 810--816 (1987)

\bibitem{Bishop2006}
Bishop, C.M.: Pattern Recognition and Machine Learning.
\newblock Springer-Verlag New York (2006).
\newblock Chapter 10: Approximate Inference

\bibitem{boyd_vandenberghe_2004}
Boyd, S., Vandenberghe, L.: Convex Optimization.
\newblock Cambridge University Press (2004).
\newblock \doi{10.1017/CBO9780511804441}

\bibitem{ChenRBD2018ode}
Chen, T.Q., Rubanova, Y., Bettencourt, J., Duvenaud, D.: Neural ordinary
  differential equations.
\newblock In: S.~Bengio, H.M. Wallach, H.~Larochelle, K.~Grauman,
  N.~Cesa{-}Bianchi, R.~Garnett (eds.) Advances in Neural Information
  Processing Systems 31: Annual Conference on Neural Information Processing
  Systems 2018, NeurIPS 2018, 3-8 December 2018, Montr{\'{e}}al, Canada, pp.
  6572--6583 (2018).
\newblock
  \urlprefix\url{http://papers.nips.cc/paper/7892-neural-ordinary-differential-equations}

\bibitem{nipsChenWZ2014}
Chen, W., Wang, Z., Zhou, J.: Large-scale {L-BFGS} using mapreduce.
\newblock In: Z.~Ghahramani, M.~Welling, C.~Cortes, N.D. Lawrence, K.Q.
  Weinberger (eds.) Advances in Neural Information Processing Systems 27:
  Annual Conference on Neural Information Processing Systems 2014, December
  8-13 2014, Montreal, Quebec, Canada, pp. 1332--1340 (2014).
\newblock
  \urlprefix\url{http://papers.nips.cc/paper/5333-large-scale-l-bfgs-using-mapreduce}

\bibitem{cherniha2017nonlinear}
Cherniha, R., Davydovych, V.: Nonlinear reaction-diffusion systems, vol. 2196.
\newblock Springer (2017).
\newblock Conditional Symmetry, Exact Solutions and their Applications in
  Biology

\bibitem{Roman2017}
Cherniha, R., Serov, M., Pliukhin, O.: Nonlinear Reaction-Diffusion-Convection
  Equations, vol. 104, 1 edn.
\newblock Chapman and Hall /CRC, New York (2017).
\newblock \doi{10.1201/9781315154848}

\bibitem{Grossmann2007}
Christian~Grossmann Hans-Gorg~Roos, M.S.: {Numerical Treatment of Partial
  Differential Equations}, 1 edn.
\newblock Springer-Verlag Berlin, Heidelberg (2007).
\newblock \doi{10.1002/0471213748}.
\newblock \urlprefix\url{https://doi.org/10.1007/978-3-540-71584-9}.
\newblock P. 23

\bibitem{deng202174}
Deng, X., Mahmoud, M.A., Yin, Q., Guo, P.: An efficient and effective deep
  convolutional kernel pseudoinverse learner with multi-filter.
\newblock Neurocomputing \textbf{457}, 74--83 (2021).
\newblock \doi{https://doi.org/10.1016/j.neucom.2021.06.041}.
\newblock
  \urlprefix\url{https://www.sciencedirect.com/science/article/pii/S0925231221009589}

\bibitem{DomnisoruMembrane22}
Domnisoru, Cristina, ans Amina~A., K., Tank, W., D.: Membrane potential
  dynamics of grid cells.
\newblock Nature \textbf{495}, 199--204 (2013)

\bibitem{Elman1990}
Elman, J.L.: Finding structure in time.
\newblock Cognitive Science \textbf{14}(2), 179--211 (1990).
\newblock \doi{10.1207/s15516709cog1402\_1}.
\newblock \urlprefix\url{https://doi.org/10.1207/s15516709cog1402\_1}

\bibitem{Feldmann2021}
Feldmann, J., Youngblood, N., Karpov, M., Gehring, H., Li, X., Stappers, M.,
  Le~Gallo, M., Fu, X., Lukashchuk, A., Raja, A.S., Liu, J., Wright, C.D.,
  Sebastian, A., Kippenberg, T.J., Pernice, W.H.P., Bhaskaran, H.: Parallel
  convolutional processing using an integrated photonic tensor core.
\newblock Nature \textbf{589}(7840), 52--58 (2021).
\newblock \doi{10.1038/s41586-020-03070-1}.
\newblock \urlprefix\url{https://doi.org/10.1038/s41586-020-03070-1}

\bibitem{FerzigerFinite28}
Ferziger, H., J., Peri\'c, M.: Finite Difference Methods.
\newblock Springer, Berlin, Heidelberg (2002).
\newblock In: Computational Methods for Fluid Dynamics.

\bibitem{Feynman2005}
Feynman, R.P.: The Principle Of Least Action in Quantum Mechanics, pp. 1--69.
\newblock WORLD SCIENTIFIC (2005).
\newblock \doi{doi:10.1142/9789812567635{\_}0001}.
\newblock \urlprefix\url{https://doi.org/10.1142/9789812567635\_0001}

\bibitem{heateq1964}
Friedman, A.: Partial differential equations of parabolic type.
\newblock Prentice-Hall, Englewood Cliffs, N.J. (1964).
\newblock \urlprefix\url{//catalog.hathitrust.org/Record/000659848}

\bibitem{quantumMe2007}
Griffiths, D.J.: Introduction to Quantum Mechanics.
\newblock Princeton, {NJ} (2007)

\bibitem{Guo1990}
Guo, P.: Numerical solution of gaussian beam propagation in nonlinear gradient
  refractive media.
\newblock Laser Technology \textbf{14}(5) (1990).
\newblock
  \urlprefix\url{http://www.cnki.com.cn/Article/CJFDTotal-JGJS199005009.htm}.
\newblock {(in Chinese)}

\bibitem{Synergetic2019b}
Guo, P.: Synergetic learning systems {(II)}: Interpretable neural network model
  with statistical physics approach.
\newblock Preprint, researchgate.net (2019).
\newblock \doi{10.13140/RG.2.2.23969.66401}.
\newblock \urlprefix\url{https://doi.org/10.13140/RG.2.2.23969.66401}.
\newblock The Fifth National Statistical Physics \& Complex Systems Conference
  (SPCSC 2019), Hefei, July 26-29, 2019

\bibitem{Guo2020archEvolve}
Guo, P.: On the structure evolutionary of the pseudoinverse learners in
  synergetic learning systems.
\newblock Preprint, researchgate.net (2020).
\newblock \doi{10.13140/RG.2.2.12262.45121}.
\newblock \urlprefix\url{https://doi.org/10.13140/RG.2.2.12262.45121}

\bibitem{SynergeticIII2020}
Guo, P.: Synergetic learning systems {(III)}: Automatic organization and
  evolution theory of neural network architecture.
\newblock Preprint, researchgate.net (2020).
\newblock \doi{10.13140/RG.2.2.23186.07360}.
\newblock \urlprefix\url{https://doi.org/10.13140/RG.2.2.23186.07360}.
\newblock The Fourth China Systems Science Conference (CSSC2020), QingDao,
  September 19-20, 2020

\bibitem{firstprinciple2020}
Guo, P.: What is the first principles for artificial intelligence (ver. 2).
\newblock Communications of the China Computer Federation \textbf{16}(10),
  53--58 (2020).
\newblock \doi{10.13140/RG.2.2.30526.72001}.
\newblock
  \urlprefix\url{https://dl.ccf.org.cn/institude/institudeDetail?\_ack=1\&id=5144072786659328}.
\newblock (in Chinese), \textcolor{blue}{ [The First Principles for Artificial
  Intelligence (ver. 1), preprint, researchgate.net,
  \url{https://doi.org/10.13140/RG.2.2.30526.72001}] }

\bibitem{Guo2023icmlc}
Guo, P.: Two-model synergetic learning systems optimization with maxwell's
  demon technique.
\newblock In: Proceedings of International Conference on Machine Learning and
  Cybernetics, {ICMLC} 2023, The University of Adelaide, Adelaide, Australia.
  9-11 July, 2023, pp. 159--164 (2023)

\bibitem{Guo1999Dynamics}
Guo, P., Awwal, A.A.S., Chen, C.L.P.: Dynamics of a coupled double-cavity
  optical interference filter.
\newblock Journal of Optics \textbf{46}(1), 167--174 (1999)

\bibitem{Guo2002}
Guo, P., Chen, C.L.P., Lyu, M.R.: Cluster number selection for a small set of
  samples using the bayesian ying-yang model.
\newblock {IEEE} Trans. Neural Networks \textbf{13}(3), 757--763 (2002).
\newblock \doi{10.1109/TNN.2002.1000144}.
\newblock \urlprefix\url{https://doi.org/10.1109/TNN.2002.1000144}

\bibitem{Guo1993Heat}
Guo, P., Chen, L., Sun, Y.G.: Laser pulse induced thermal distribution in
  interference filter.
\newblock Journal of Beijing Normal University (Natural Science Edition)
  \textbf{29}(1), 82--86 (1993).
\newblock
  \urlprefix\url{http://www.cnki.com.cn/Article/CJFDTOTAL-BSDZ199301018.htm}.
\newblock {(in Chinese)}, Received: 1992-09-03

\bibitem{method-SLS}
Guo, P., Hou, J.X., Zhao, B.: Methodology for building synergetic learning
  system.
\newblock Preprint, researchget.net, (ICCS 2020, English version) (2019).
\newblock \doi{10.13140/RG.2.2.10146.07368}.
\newblock \urlprefix\url{http://dx.doi.org/10.13140/RG.2.2.10146.07368}.
\newblock The third China Systems Science Conference (CSSC2019), Changsha, May
  18-19, 2019. (Chinese version)

\bibitem{Guo2001OMLNN}
Guo, P., Lyu, M.R.: A new approach to optical multilayer learning neural
  network.
\newblock In: H.R. Arabnia (ed.) Proceedings of International Conference on
  artificial intelligence {(IC-AI'01)}, vol.~I, pp. 426--432. CSREA Press, Las
  Vegas, Nevada, USA (2001).
\newblock
  \urlprefix\url{https://www.researchgate.net/publication/331463645\_A\_New\_Approach\_to\_Optical\_Multilayer\_Learning\_Neural\_Network}

\bibitem{Guo2003reg}
Guo, P., Lyu, M.R., Chen, C.L.P.: Regularization parameter estimation for
  feedforward neural networks.
\newblock {IEEE} Trans. Systems, Man, and Cybernetics, Part {B} \textbf{33}(1),
  35--44 (2003).
\newblock \doi{10.1109/TSMCB.2003.808176}.
\newblock \urlprefix\url{https://doi.org/10.1109/TSMCB.2003.808176}

\bibitem{Guo1990AOS}
Guo, P., Sun, Y.G.: Gaussian beam propagation with nonlinear medium limiter.
\newblock Acta Optica Sinica \textbf{10}(12) (1990).
\newblock
  \urlprefix\url{http://www.cnki.com.cn/Article/CJFDTOTAL-GXXB199012006.htm}.
\newblock (in Chinese)

\bibitem{Guo1993HeatDyna}
Guo, P., Sun, Y.G.: Thermo--optic dynamics analysis of interference filter
  bistable device.
\newblock Journal of Beijing Normal University (Natural Science Edition)
  \textbf{29}(2), 189--193 (1993).
\newblock
  \urlprefix\url{http://www.cnki.com.cn/Article/CJFDTOTAL-BSDZ199302009.htm}.
\newblock {(in Chinese)}, Received: 1992-11-03

\bibitem{Guo1987}
Guo, P., Sun, Y.G., Xiong, J., Wang, W., Jiang, Z.: Nonlinear three-mirror
  {Fabry-Perot} multistability element, pp. 83--87.
\newblock World Scientific, Singapore (1987).
\newblock
  \urlprefix\url{https://www.researchgate.net/publication/330955256\_Nonlinear\_three-mirror\_Fabry-Perot\_multistability\_element}

\bibitem{Synergetic2019}
Guo, P., Yin, Q.: Synergetic learning systems: Concept, architecture, and
  algorithms (2020).
\newblock \urlprefix\url{https://arxiv.org/abs/2006.06367}

\bibitem{GuoYZ2015image}
Guo, P., Yin, Q., Zhou, X.L.: Image Semantic Analysis.
\newblock Science Press, Beijing (2015).
\newblock \urlprefix\url{http://www.ecsponline.com/goods.php?id=174834}.
\newblock (in Chinese)

\bibitem{WeinanPNAS2018}
Han, J., Jentzen, A., E, W.: Solving high-dimensional partial differential
  equations using deep learning.
\newblock Proceedings of the National Academy of Sciences \textbf{115}(34),
  8505--8510 (2018).
\newblock \doi{10.1073/pnas.1718942115}.
\newblock \urlprefix\url{https://www.pnas.org/content/115/34/8505}

\bibitem{HanJCP2019}
Han, J., Zhang, L., E, W.: Solving many-electron schr{\"o}dinger equation using
  deep neural networks.
\newblock Journal of Computational Physics \textbf{399}, 108929 (2019).
\newblock \doi{10.1016/j.jcp.2019.108929}

\bibitem{eccvHeZRS2016ResNet2}
He, K., Zhang, X., Ren, S., Sun, J.: Identity mappings in deep residual
  networks.
\newblock In: Computer Vision - {ECCV} 2016 - 14th European Conference,
  Amsterdam, The Netherlands, October 11-14, 2016, Proceedings, Part {IV}, pp.
  630--645 (2016).
\newblock \doi{10.1007/978-3-319-46493-0\_38}.
\newblock \urlprefix\url{https://doi.org/10.1007/978-3-319-46493-0\_38}

\bibitem{Hinton2006}
Hinton, G.E., Salakhutdinov, R.R.: Reducing the dimensionality of data with
  neural networks.
\newblock Science \textbf{313}, 504--507 (2006)

\bibitem{Hochreiter1997LSTM}
Hochreiter, S., Schmidhuber, J.: {Long Short-Term Memory}.
\newblock Neural Computation \textbf{9}(8), 1735--1780 (1997).
\newblock \doi{10.1162/neco.1997.9.8.1735}.
\newblock \urlprefix\url{https://doi.org/10.1162/neco.1997.9.8.1735}

\bibitem{Hughes2019}
Hughes, T.W., Williamson, I.A.D., Minkov, M., Fan, S.: Wave physics as an
  analog recurrent neural network.
\newblock Science Advances \textbf{5}(12), eaay6946 (2019).
\newblock \doi{10.1126/sciadv.aay6946}.
\newblock \urlprefix\url{https://advances.sciencemag.org/content/5/12/eaay6946}

\bibitem{numericalRDE1}
Iliev, O.P., Margenov, S.D., Minev, P.D., Vassilevski, P.S., Zikatanov, L.T.
  (eds.): Numerical Solution of Partial Differential Equations: Theory,
  Algorithms, and Their Applications, \emph{Springer Proceedings in Mathematics
  \& Statistics}, vol.~45, 1 edn.
\newblock Springer-Verlag, New York (2013)

\bibitem{kingma2014adam}
Kingma, D., Ba, J.: {ADAM}: A method for stochastic optimization.
\newblock In: 3rd International Conference on Learning Representations, {ICLR}
  2015, San Diego, CA, USA, May 7-9, 2015, Conference Track Proceedings,
  https://iclr.cc/archive/www/doku.php\%3Fid=iclr2015:accepted-main.html
  (2014).
\newblock \urlprefix\url{https://arxiv.org/abs/1412.6980}

\bibitem{numericalRDE}
Lapidus, L.: Numerical Solution of Partial Differential Equations in Science
  and Engineering.
\newblock Wiley-Interscience (1999)

\bibitem{LeCun2018jicai}
LeCun, Y.: Learning world models: the next step towards {AI}.
\newblock Keynote, the 27th {IJCAI} (2018)

\bibitem{Deeplearning}
LeCun, Y., Bengio, Y., Hinton, G.: Deep learning.
\newblock Nature \textbf{521}(7553), 436--444 (2015).
\newblock \doi{10.1038/nature14539}

\bibitem{Lecun2006A13}
Lecun, Y., Chopra, S., Hadsell, R., Ranzato, M., Huang, F.J.: A tutorial on
  energy-based learning.
\newblock In: Predicting Structured Data. MIT Press (2006)

\bibitem{TrendsPDE2014}
Leugering, G., Benner, P., Engell, S., Griewank, A., Harbrecht, H., Hinze, M.,
  Rannacher, R., Ulbrich, S. (eds.): Trends in {PDE} Constrained Optimization,
  \emph{International Series of Numerical Mathematics}, vol. 165, 4 edn.
\newblock Birkh\"auser, Cham (2014).
\newblock \doi{10.1007/978-3-319-05083-6}.
\newblock \urlprefix\url{https://doi.org/10.1007/978-3-319-05083-6}

\bibitem{Lipton2015RNN}
Lipton, Z.C.: A critical review of recurrent neural networks for sequence
  learning (2015).
\newblock \urlprefix\url{http://arxiv.org/abs/1506.00019}

\bibitem{Liu1989BGFS}
Liu, D.C., Nocedal, J.: On the limited memory bfgs method for large scale
  optimization.
\newblock Mathematical Programming \textbf{45}(1), 503--528 (1989).
\newblock \doi{10.1007/BF01589116}.
\newblock \urlprefix\url{https://doi.org/10.1007/BF01589116}

\bibitem{LongLD2019PDEnet}
Long, Z., Lu, Y., Dong, B.: {PDE-Net} 2.0: Learning {PDEs} from data with a
  numeric-symbolic hybrid deep network.
\newblock Journal of Computational Physics \textbf{399}, 108925 (2019).
\newblock \doi{https://doi.org/10.1016/j.jcp.2019.108925}.
\newblock
  \urlprefix\url{http://www.sciencedirect.com/science/article/pii/S0021999119306308}

\bibitem{LuZLD2018icml}
Lu, Y., Zhong, A., Li, Q., Dong, B.: Beyond finite layer neural networks:
  Bridging deep architectures and numerical differential equations.
\newblock In: J.G. Dy, A.~Krause (eds.) Proceedings of the 35th International
  Conference on Machine Learning, {ICML} 2018, Stockholmsm{\"{a}}ssan,
  Stockholm, Sweden, July 10-15, 2018, \emph{Proceedings of Machine Learning
  Research}, vol.~80, pp. 3282--3291. {PMLR} (2018).
\newblock \urlprefix\url{http://proceedings.mlr.press/v80/lu18d.html}

\bibitem{Miriyev2020}
Miriyev, A., Kova{\v c}, M.: Skills for physical artificial intelligence.
\newblock Nature Machine Intelligence \textbf{2}(11), 658--660 (2020).
\newblock \doi{10.1038/s42256-020-00258-y}.
\newblock \urlprefix\url{https://doi.org/10.1038/s42256-020-00258-y}

\bibitem{Neal1996Infinite}
Neal, R.M.: Priors for Infinite Networks, pp. 29--53.
\newblock Springer New York, New York, NY (1996).
\newblock \doi{10.1007/978-1-4612-0745-0{\_}2}.
\newblock \urlprefix\url{https://doi.org/10.1007/978-1-4612-0745-0\_2}.
\newblock In Bayesian Learning for Neural Networks

\bibitem{Self-Organization26}
Nikolis, G., Prigogine, I.: Self-Organization in Non-Equilibrium Systems.
\newblock Wiley, New York (1977)

\bibitem{Niu2008Improving32}
Niu, H., Guo, P., Ji, L., Zhao, Q., Jiang, T.: Improving image quality of
  diffuse optical tomography with a projection-error-based adaptive
  regularization method.
\newblock Optics Express \textbf{16}(17), 12423--34 (2008)

\bibitem{PaoCV1992nonlinear}
Pao, C.V.: Nonlinear parabolic and elliptic equations.
\newblock Springer US, Boston, MA (1992).
\newblock \doi{10.1007/978-1-4615-3034-3}.
\newblock \urlprefix\url{https://doi.org/10.1007/978-1-4615-3034-3}

\bibitem{Pearson1993}
Pearson, J.E.: Complex patterns in a simple system.
\newblock Science \textbf{261}(5118), 189--192 (1993).
\newblock \urlprefix\url{http://dx.doi.org/10.1126/science.261.5118.189}

\bibitem{PhysRevResearch2020}
Pfau, D., Spencer, J.S., Matthews, A.G.D.G., Foulkes, W.M.C.: Ab initio
  solution of the many-electron schr\"odinger equation with deep neural
  networks.
\newblock Phys. Rev. Research \textbf{2}, 033429 (2020).
\newblock \doi{10.1103/PhysRevResearch.2.033429}.
\newblock
  \urlprefix\url{https://link.aps.org/doi/10.1103/PhysRevResearch.2.033429}

\bibitem{HopfieldNAYN2020}
Ramsauer, H., Sch{\"{a}}fl, B., Lehner, J., Seidl, P., Widrich, M., Gruber, L.,
  Holzleitner, M., Pavlovic, M., Sandve, G.K., Greiff, V., Kreil, D.P., Kopp,
  M., Klambauer, G., Brandstetter, J., Hochreiter, S.: Hopfield networks is all
  you need.
\newblock CoRR \textbf{abs/2008.02217} (2020).
\newblock \urlprefix\url{https://arxiv.org/abs/2008.02217}

\bibitem{Fokker-PlanckEq1996}
Risken, H.: {The Fokker-Planck Equation: Methods of Solution and Applications},
  2 edn.
\newblock Springer Series in Synergetics. Springer-Verlag Berlin Heidelberg
  (1996).
\newblock \doi{https://doi.org/10.1007/978-3-642-61544-3}.
\newblock EBook Published: 06 December 2012.

\bibitem{photonics2019}
Saleh, B.E.A., Teich, M.C.: Fundamentals of Photonics, 3 edn.
\newblock John Wiley \& Sons, Inc, Hoboken, NJ (2019).
\newblock \doi{10.1002/0471213748}.
\newblock
  \urlprefix\url{https://onlinelibrary.wiley.com/doi/book/10.1002/0471213748}

\bibitem{Sanchez2018}
Sanchez-Lengeling, B., Aspuru-Guzik, A.: Inverse molecular design using machine
  learning: Generative models for matter engineering.
\newblock Science \textbf{361}(6400), 360--365 (2018).
\newblock \doi{10.1126/science.aat2663}.
\newblock \urlprefix\url{https://science.sciencemag.org/content/361/6400/360}

\bibitem{sohl2015deep}
Sohl-Dickstein, J., Weiss, E.A., Maheswaranathan, N., Ganguli, S.: Deep
  unsupervised learning using nonequilibrium thermodynamics.
\newblock In: F.R. Bach, D.M. Blei (eds.) Proceedings of the 32nd International
  Conference on Machine Learning, {ICML} 2015, Lille, France, 6-11 July 2015,
  \emph{{JMLR} Workshop and Conference Proceedings}, vol.~37, pp. 2256--2265.
  JMLR.org (2015).
\newblock
  \urlprefix\url{http://proceedings.mlr.press/v37/sohl-dickstein15.html}

\bibitem{numericalAN}
Stoer, J., Bulirsch, R.: Introduction to Numerical Analysis, \emph{Texts in
  Applied Mathematics}, vol.~12, 3 edn.
\newblock Springer-Verlag New York (2002).
\newblock \doi{10.1007/978-0-387-21738-3}.
\newblock P.243, Translated by Gautschi, W., Bartels, R., Witzgall, C.

\bibitem{TYCBSun2020}
{Sun}, Y., {Xue}, B., {Zhang}, M., {Yen}, G.G., {Lv}, J.: Automatically
  designing cnn architectures using the genetic algorithm for image
  classification.
\newblock IEEE Transactions on Cybernetics \textbf{50}(9), 3840--3854 (2020).
\newblock \doi{10.1109/TCYB.2020.2983860}.
\newblock \urlprefix\url{https://doi.org/10.1109/TCYB.2020.2983860}

\bibitem{Sun1987}
Sun, Y.G., Jiang, Z., Xiong, J., Wang, W., Guo, P., Shang, S.: Optical
  Bistability in Very Thin {ZnSe} Film, pp. 95--97.
\newblock World Scientific, Singapore (1987)

\bibitem{Svelto1974}
Svelto, O.: Self-Focusing, Self-Trapping, and Self-Phase Modulation of Laser
  Beams, vol.~12, pp. 1--51.
\newblock Elsevier (1974).
\newblock \doi{https://doi.org/10.1016/S0079-6638(08)70263-4}.
\newblock
  \urlprefix\url{http://www.sciencedirect.com/science/article/pii/S0079663808702634}

\bibitem{Tikhonov1977}
Tikhonov, A.N., Arsenin, V.Y.: Solutions of Ill-Posed Problems.
\newblock Wiley, Hoboken \& New York (1977).
\newblock Chapter 2

\bibitem{Turing1952}
Turing, A.M.: The chemical basis of morphogenesis.
\newblock Philosophical Transactions of the Royal Society of London, Series B
  \textbf{237}(641), 37--72 (1952)

\bibitem{nipsVaswaniSPUJGKP2017}
Vaswani, A., Shazeer, N., Parmar, N., Uszkoreit, J., Jones, L., Gomez, A.N.,
  Kaiser, L., Polosukhin, I.: Attention is all you need.
\newblock In: I.~Guyon, U.~von Luxburg, S.~Bengio, H.M. Wallach, R.~Fergus,
  S.V.N. Vishwanathan, R.~Garnett (eds.) Advances in Neural Information
  Processing Systems 30: Annual Conference on Neural Information Processing
  Systems 2017, 4-9 December 2017, Long Beach, CA, {USA}, pp. 5998--6008
  (2017).
\newblock
  \urlprefix\url{http://papers.nips.cc/paper/7181-attention-is-all-you-need}

\bibitem{Wang2019TCYB}
{Wang}, J.W., {Wu}, H.N.: Design of suboptimal local piecewise fuzzy controller
  with multiple constraints for quasi-linear spatiotemporal dynamic systems.
\newblock IEEE Transactions on Cybernetics pp. 1--13 (2019).
\newblock \doi{10.1109/TCYB.2019.2923461}

\bibitem{Wang1987}
Wang, W., Sun, Y.G., Xiong, J., Jiang, Z., Guo, P., Shang, S.: Low-power
  Optical Bistability in Improved {ZnSe} Interference Filters, pp. 57--59.
\newblock World Scientific, Singapore (1987).
\newblock
  \urlprefix\url{https://www.researchgate.net/publication/348390364\_Low-power\_Optical\_Bistability\_in\_Improved\_ZnSe\_Interference\_Filters}

\bibitem{Wells2008Elliptic}
Wells, R.O.: Elliptic Operator Theory, pp. 108--153.
\newblock Springer New York, New York, NY (2008).
\newblock \doi{10.1007/978-0-387-73892-5{\_}4}.
\newblock \urlprefix\url{https://doi.org/10.1007/978-0-387-73892-5\_4}

\bibitem{Wick1954}
Wick, G.C.: Properties of bethe-salpeter wave functions.
\newblock Phys. Rev. \textbf{96}, 1124--1134 (1954).
\newblock \doi{10.1103/PhysRev.96.1124}.
\newblock \urlprefix\url{https://link.aps.org/doi/10.1103/PhysRev.96.1124}

\bibitem{Willems1972}
Willems, J.C.: Dissipative dynamical systems part {I}: General theory.
\newblock Archive for Rational Mechanics and Analysis \textbf{45}(5), 321--351
  (1972).
\newblock \doi{10.1007/BF00276493}.
\newblock \urlprefix\url{https://doi.org/10.1007/BF00276493}

\bibitem{Wolfram2002}
Wolfram, S.: A New Kind of Science.
\newblock Wolfram Media, Champaign, IL (2002).
\newblock \urlprefix\url{https://www.wolframscience.com}

\bibitem{light2021}
Wu, H., Dai, Q.: Artificial intelligence accelerated by light.
\newblock Nature \textbf{589}, 25--26 (2021).
\newblock \doi{10.1038/d41586-020-03572-y}.
\newblock \urlprefix\url{https://doi.org/10.1038/d41586-020-03572-y}.
\newblock News \& Views

\bibitem{Xu2021ONN}
Xu, X., Tan, M., Corcoran, B., Wu, J., Boes, A., Nguyen, T.G., Chu, S.T.,
  Little, B.E., Hicks, D.G., Morandotti, R., Mitchell, A., Moss, D.J.: 11
  {TOPS} photonic convolutional accelerator for optical neural networks.
\newblock Nature \textbf{589}(7840), 44--51 (2021).
\newblock \doi{10.1038/s41586-020-03063-0}.
\newblock \urlprefix\url{https://doi.org/10.1038/s41586-020-03063-0}

\bibitem{Qixiao16}
Ye, Q.X.: Introduction to reaction diffusion equation.
\newblock Practice and understanding of mathematics pp. 48--56 (1984-02).
\newblock In Chinese

\bibitem{zhangyi2023nmODE}
Yi, Z.: nmode: neural memory ordinary differential equation.
\newblock Artificial Intelligence Review \textbf{56}(12), 14403--14438 (2023).
\newblock \doi{10.1007/s10462-023-10496-2}.
\newblock \urlprefix\url{https://doi.org/10.1007/s10462-023-10496-2}

\bibitem{BPIL2020}
Yin, Q., Xu, B., Zhou, K., Guo, P.: Bayesian pseudoinverse learners: From
  uncertainty to deterministic learning.
\newblock IEEE Transactions on Cybernetics pp. 1--12 (2021).
\newblock \doi{10.1109/TCYB.2021.3079906}

\bibitem{TYCBZheng2020}
{Zheng}, N., {Du}, S., {Wang}, J., {Zhang}, H., {Cui}, W., {Kang}, Z., {Yang},
  T., {Lou}, B., {Chi}, Y., {Long}, H., {Ma}, M., {Yuan}, Q., {Zhang}, S.,
  {Zhang}, D., {Ye}, F., {Xin}, J.: Predicting covid-19 in china using hybrid
  ai model.
\newblock IEEE Transactions on Cybernetics \textbf{50}(7), 2891--2904 (2020).
\newblock \doi{10.1109/TCYB.2020.2990162}.
\newblock \urlprefix\url{https://doi.org/10.1109/TCYB.2020.2990162}

\bibitem{Zienkiewicz2013}
Zienkiewicz, O.C., Taylor, R.L., Zhu, J.Z.: The Finite Element Method: Its
  Basis and Fundamentals, p. 756.
\newblock Butterworth-Heinemann, Oxford (2013).
\newblock \doi{10.1016/B978-1-85617-633-0.00019-8}.
\newblock \urlprefix\url{https://doi.org/10.1016/B978-1-85617-633-0.00019-8}

\end{thebibliography}

\end{document}